\documentclass[twoside,11pt]{article}

\usepackage{blindtext}

\usepackage{jmlr2e}
\usepackage{multirow}
\usepackage{multicol}
\usepackage{float}
\usepackage{wrapfig}
\usepackage{diagbox}
\usepackage{comment}
\usepackage{amsmath}

\usepackage{lastpage}

\ShortHeadings{\parbox{\textwidth}{\centering Analyzing the Effect of Embedding Norms and Singular Values to Oversmoothing in Graph Neural Networks}}{Kelesis et  al.}
\firstpageno{1}

\begin{document}

\title{Analyzing the Effect of Embedding Norms and Singular Values to Oversmoothing in Graph Neural Networks}

\author{\name Dimitrios Kelesis \email dkelesis@iit.demokritos.gr\\
       \addr School of Electrical Engineering and Computer Science\\
       National Technical University of Athens\\
       Athens, Heroon Polytechniou 9, Greece\\
       \&\\
       \addr Institute of Informatics and Telecommunications\\
       National Centre for Scientific Research ``Demokritos"\\
       Athens, Patr. Gregoriou E \& 27 Neapoleos, Greece
       \AND
       \name Dimitris Fotakis \email fotakis@cs.ntua.gr\\
       \addr School of Electrical Engineering and Computer Science\\
       National Technical University of Athens\\
       Athens, Heroon Polytechniou 9, Greece\\
       \&\\
       Archimedes Research Unit, Athena RC\\
       Athens, Artemidos 1, Greece
       \AND
       \name Georgios Paliouras \email paliourg@iit.demokritos.gr\\
       \addr Institute of Informatics and Telecommunications\\
       National Centre for Scientific Research ``Demokritos"\\
       Athens, Patr. Gregoriou E \& 27 Neapoleos, Greece}


\maketitle

\begin{abstract}
In this paper, we study the factors that contribute to the effect of oversmoothing in deep Graph Neural Networks (GNNs). Specifically, our analysis is based on a new metric (Mean Average Squared Distance - $MASED$) to quantify the extent of oversmoothing. We derive layer-wise bounds on $MASED$, which aggregate to yield global upper and lower distance bounds. Based on this quantification of oversmoothing, we further analyze the importance of two different properties of the model; namely the norms of the generated node embeddings, along with the largest and smallest singular values of the weight matrices.\\  
Building on the insights drawn from the theoretical analysis, we show that oversmoothing increases as the number of trainable weight matrices and the number of adjacency matrices increases. We also use the derived layer-wise bounds on $MASED$ to form a proposal for decoupling the number of hops (i.e., adjacency depth) from the number of weight matrices. In particular, we introduce \emph{G-Reg}, a regularization scheme that increases the bounds, and demonstrate through extensive experiments that by doing so node classification accuracy increases, achieving  robustness at large depths.\\ 
We further show that by reducing oversmoothing in deep networks, we can achieve better results in some tasks than using shallow ones. Specifically, we experiment with a ``cold start" scenario, i.e., when there is no feature information for the unlabeled nodes. Finally, we show empirically the trade-off between receptive field size (i.e., number of weight matrices) and performance, using the $MASED$ bounds. This is achieved by distributing adjacency hops across a small number of trainable layers, avoiding the extremes of under- or over-parameterization of the GNN. 
\end{abstract}


\section{Introduction}
Graph Neural Networks (GNNs) have proven very effective for analyzing and learning from graph-structured data. Their ability to combine node attributes with the inherent connectivity of graphs has led to remarkable advances in many fields, including social network analysis, bioinformatics, and recommendation systems. Various problems in these domains can be modeled as node classification \citep{Zhang,graphsage}, link prediction \citep{link_pred1,link_pred2} and graph classification \citep{chemistry} tasks, and addressed using GNNs. GNNs work by iteratively aggregating information from a node’s neighbors, which allows them to learn representations that capture both local details and global structure.\\
Despite their success, one of the most significant challenges in extending GNNs to deeper architectures is the phenomenon known as oversmoothing. As more layers are added, the repeated process of neighbor aggregation tends to make node representations increasingly similar, eventually leading to a loss of unique, distinguishing features. This effect undermines the performance of deep GNN models, as they become less able to differentiate between nodes with different characteristics \citep{A_note, Li_et, JK_Nets, Suzuki}.\\
To counter oversmoothing, researchers have explored a number of strategies that modify the way information is propagated through the network. One family of methods modify the structure of the graph or network's architecture to reduce the effect of information aggregation. Such methods include the removal of edges or nodes from the graph \citep{Dropedge, Dropnode}, and the introduction of residual and skip connections \citep{GCNII, JK_Nets, deep_gcn}. The latter allow original node features to bypass one or more layers, thereby preserving important details across the network. Other approaches, such as the Approximate Personalized Propagation of Neural Predictions (APPNP) \citep{APPNP} and Simplified Graph Convolution (SGC) \citep{sgc}, adjust the propagation mechanism so that the aggregation process is decoupled from feature transformation. APPNP, for example, uses a personalized PageRank scheme to balance local and global information, while SGC simplifies the network by collapsing multiple layers into a single transformation. More recent strategies explore the impact of the activation functions \citep{tanh_over, Kelesis} or leverage the Dirichlet energy \citep{dirichlet_energy} to reduce oversmoothing \citep{Dropedge, JK_Nets, dirichlet_energy}. All of the above-mentioned efforts address under conditions the problem of oversmoothing, but fail to provide a general solution. Therefore, there is still the need for a deeper understanding of the problem, which will open the possibilities for novel and effective solutions.\\
An important tool in our understanding of the problem of oversmoothing is a quantitative measure of oversmoothing in a given model. Building on prior studies \citep{mads}, we propose a novel distance measure, namely $MASED$, and conduct an in-depth investigation into its properties. We derive upper and lower bounds on its value and examine how the lower bound can be increased, in order to maintain node embedding variance and reduce oversmoothing. In this direction, we highlight the key role of the node embedding norms and of the smallest singular values of weight matrices, which have been largely overlooked. Additionally, we shed light on the relationship between the number of adjacency hops and weight matrices in GNNs. Our analysis provides insights into the characteristics of a deep model that lead to oversmoothing and how this, in turn, affects model performance. These findings can motivate the design of more effective strategies for mitigating the impact of oversmoothing in deep GNNs. We also propose one such mitigation strategy, in the form of a novel regularization method, named \emph{G-Reg}, which aims to reduce co-linearity between the rows of the weight matrices and, in turn, increase their smallest singular values. \\
In summary, the main contributions of this work are as follows:\\ 
\noindent \textbullet \hspace{.01cm} \textbf{Contribution on Oversmoothing Quantification:} We introduce the Mean Average Squared Euclidean Distance ($MASED$) to quantify oversmoothing in GNNs. We derive the upper and lower bounds of $MASED$ and reveal the relationship between the distances among node representations and the structure of the weight matrix, highlighting how the independence of its rows affects oversmoothing. Moreover, we show that $MASED$ values can predict oversmoothing early in the learning process. A rapid decline, accompanied by a reduction in the weight matrix's singular values, leads to poor model performance.\\ 
\noindent \textbullet \hspace{.01cm} \textbf{The effect of embedding norms and angles on oversmoothing:} We measure the average angle between the centroids (i.e., average embedding of nodes belonging to the same class) of the training nodes and the average norms of node embeddings. We observe that if norms are close to zero then the model is incapable of learning. However, if the norms are large enough and the model has the needed capacity, small angles suffice to maintain competitive performance.\\
\noindent \textbullet \hspace{.01cm} \textbf{New weight regularization method (\emph{G-Reg}):} We propose a regularization of the values of the weight matrix in order to maintain higher values of $MASED$ and higher values of the embedding norms. We experimentally confirm the benefits of \emph{G-Reg}, utilizing deep GCNs, residual GCNs, and SGCs, with up to 32 layers across 7 node classification tasks. We show that the proposed regularization reduces oversmoothing, and demonstrate its benefits, in the presence of the “cold start” situation, where node features are available only for the labeled nodes.\\
\noindent \textbullet \hspace{.01cm} \textbf{Reducing weight redundancy in multi-hop aggregation:} We propose using fewer weight matrices than the number of neighborhood hops, improving learnability and reducing oversmoothing.

\section{Notations and Preliminaries}
\subsection{Notations}
We focus on the common task of node classification on a graph. The graph under investigation is G($\mathbb{V,E}$,X), with $|\mathbb{V}| = N$ nodes $u_i \in \mathbb{V}$, edges $(u_i, u_j) \in \mathbb{E}$ and $X=[x_1,...,x_N]^T \in \mathbb{R}^{N \times C}$ the initial node features. The edges form an adjacency matrix $A \in \mathbb{R}^{N \times N}$ where edge $(u_i, u_j)$ is associated with element $A_{i,j}$. $A_{i,j}$ can take arbitrary real values indicating the weight (strength) of edge $(u_i, u_j)$. Node degrees are represented through a diagonal matrix $D \in \mathbb{R}^{N \times N}$, where each element $d_i$ represents the sum of edge weights connected to node i. During training, only the labels of a subset $V_l \in \mathbb{V}$ are available. The task is to learn a node classifier, that predicts the label of each node using the graph topology and the given feature vectors.\\
Graph Convolution Network (GCN), originally proposed by \cite{Kipf_gcn}, utilizes a feed-forward propagation as:
\begin{equation}\label{eq:gcn}
    H^{(l+1)} = \sigma (\hat{A}H^{(l)}W^{(l)}),
\end{equation}
where $H^{(l)} = [h^{(l)}_1,...,h^{(l)}_N]$ are the node representations (or embeddings) of the $l$-th layer, with $h^{(l)}_i$ standing for the representation of node i; $\hat{A} = \hat{D}^{-1/2} (A+I)\hat{D}^{-1/2}$ denotes the augmented symmetrically normalized adjacency matrix after self-loop addition, with $\hat{D}$ corresponding to the degree matrix of $A+I$; $\sigma(\cdot)$ is a nonlinear element-wise function, i.e. the activation function, typically a ReLU; and $W^{(l)} \in \mathbb{R}^{d \times d}$ is the trainable weight matrix of the $l$-th layer, with $d$ being the hidden size of the model, i.e., model width.\\

\noindent Residual GCN (ResGCN) \citep{resgcn} enhances the standard GCN by introducing skip (residual) connections. These connections allow the original node features to be directly added to the output of a layer, which helps to preserve crucial information that might otherwise be lost during deep aggregation. This modification can be expressed as:
\begin{equation}
    H^{(l+1)} = \sigma \left( \hat{A} H^{(l)} W^{(l)} + H^{(0)}\right),
\end{equation}
where the addition of $H^{(0)}$ ($H^{(0)} = X$) allows the initial features to be used directly at layer $l$.\\
The Simplified Graph Convolution (SGC) model aims to reduce the complexity of deep GCNs, by removing non-linearities and collapsing multiple propagation layers into a single linear transformation. In SGC, the aggregated node features are pre-computed by repeatedly applying the normalized adjacency matrix to the input features:
\begin{equation}
    H^{(L)} = softmax(\hat{A}^L X W),
\end{equation}
where $K$ is the number of propagation steps \citep{sgc}.

\subsection{Understanding Oversmoothing}
\citet{Li_et} demonstrated that graph convolution is a type of Laplacian smoothing, which GNNs utilize to generate node representations that are homogeneous within each graph cluster, enhancing performance on semi-supervised tasks. However, stacking multiple layers results in repeated smoothing operations, leading to oversmoothing, where node representations become too similar and much of the initial information is lost.\\
\citet{Suzuki} generalized the idea in \citet{Li_et}, considering also that the ReLU activation function maps to a positive cone of the space of trainable node embeddings. They characterize oversmoothing as convergence of node embeddings to a limited subspace and provide an estimate of the speed of convergence to this subspace, i.e., the rate at which the distance of node representations from the oversmoothing subspace $M$ decreases as depth increases (details can be found in \citep{Suzuki}).\\
According to \citet{Suzuki}, deep GCNs are susceptible to oversmoothing, with the model's only defense against this effect being the product of the largest singular values of the weight matrices. This multiplication arises from the 1-Lipschitzness of ReLU, combined with the propagation scheme of GCN (Equation \ref{eq:gcn}). The intuition behind this result can be seen if we remove the activation function from Equation \ref{eq:gcn}, obtaining the final node representations by $H^{(L)}=A^LX\prod\limits_{i=1}^L{W^{(i)}}$. Therefore, the product of the largest singular values of the weight matrices serves as the upper bound of the node representations. Theorem \ref{theo:suzuki} states this relationship more formally.
\begin{theorem}[\citet{Suzuki}]\label{theo:suzuki}
Let $s_l=\prod\limits_{h=1}^{L}{s_{lh}}$ where $s_{lh}$ is the largest singular value of weight matrix $W_{lh}$ and s = $sup_{l\in N^+} s_{l}$. Then the distance from the oversmoothing subspace $M$ is measured as follows: $d_M(X^{(l)}) = O((s\lambda)^l)$, where l is
the layer number, $\lambda$ is the smallest non-zero eigenvalue of $I - \hat{A}$, and if $s\lambda < 1$ the distance from the oversmoothing subspace ($d_M$) exponentially approaches
zero.
\end{theorem}

\section{Related Work}
\subsection{Oversmoothing reduction}
Research on GNNs has revealed that increasing the depth of the network often leads to oversmoothing. As additional layers are stacked, the iterative process of aggregating information from neighboring nodes causes their representations to converge, eventually making nodes indistinguishable. This loss of diversity in node features detrimentally affects the ability of GNNs to perform tasks such as node classification, as the model’s capacity to capture unique structural and feature-related information is compromised \citep{Li_et, Suzuki}. Several studies have analyzed and confirmed the inevitability of oversmoothing with increasing depth, emphasizing that deeper architectures, while theoretically promising for capturing global patterns, risk degrading performance due to the flattening of the learned representations \citep{A_note, benefits_from_deep}.\\
\noindent Two main approaches to counteract oversmoothing appear in the literature. One approach alters the graph topology to slow down the message passing process. For example, techniques such as DropEdge and DropNode randomly remove edges or nodes from the graph, thereby reducing the rate at which the smoothing effect propagates across the network \citep{Dropedge,Dropnode}. Another category of methods targets the GNN model architecture or the training process dynamics. The goal of this approach is to retain the inherent benefits of deep architectures while preventing over-homogenization of node features. Such strategies include the use of skip or residual connections, which have been mentioned in previous sections, as well as methods that modify the propagation mechanics to decouple feature transformation from information aggregation \citep{APPNP, GCNII}. Beyond these two main approaches, other methods introduce specialized weight initialization methods \citep{gnn_init_1, Kelesis_init} or modified activation functions \citep{tanh_over}. All these approaches provide valuable insights into how oversmoothing can be reduced, but each comes with its own trade-offs in terms of complexity and potential effects on model performance.
 
\subsection{Mean Average Distance (MAD)}
Mean Average Distance (MAD) has been proposed by \cite{mads} as a proxy for measuring oversmoothing in Graph Neural Networks. By computing the average distance between node embeddings within a graph, MAD provides an intuitive measure of how diverse the node representations are. In well-performing networks, node embeddings should remain distinct enough to capture important differences in the data, resulting in higher MAD values. In contrast, when oversmoothing occurs, successive layers of neighborhood aggregation force the node representations to converge towards a similar value, which is reflected in lower MAD values. This reduction in the mean distance serves as an indicator that the network is losing the necessary variance among node features, which is critical for robust downstream tasks \citep{mads}.\\
Mean Average Distance (MAD) can be computed using various distance metrics. \citet{mads} opted for a cosine distance, which focuses on the angular difference between vectors. While cosine similarity has proven useful for capturing relational trends between node representations, it may overlook absolute differences in embedding magnitudes, potentially limiting the granularity of the insights it provides into the oversmoothing phenomenon \citep{cos_drawbacks}.\\
\noindent In our work, we use the squared Euclidean distance in the calculation of MAD (termed as the $MASED$ metric), in order to capture the absolute magnitude differences between node embeddings. This change enables us to extract clearer explanations about the impact of the weight matrix on node representation collapse, and offers more actionable insights for reducing oversmoothing in deep GNN architectures.

\section{Theoretical Analysis}
\subsection{Mean Average Squared Euclidean Distance (MASED)}
The Mean Average Squared Euclidean Distance ($MASED$) of node representations can act as a surrogate to measure the extent of oversmoothing in node representations, while allowing a rigorous analysis of its properties and a derivation of the connection with weight matrix properties. $MASED$ is also highly valuable for capturing the dynamic behavior of a GNN throughout its training process. During the early training epochs, slight variations in the model’s learning dynamics can be detected through changes in the average distances among node embeddings. A rapid decrease in $MASED$ indicates that the model quickly begins to enforce uniformity on the node representations, serving as an early alarm for the onset of oversmoothing even before the overall performance declines. This sensitivity makes $MASED$ a robust proxy that reflects subtle shifts in the network, often correlating with other indicators, such as shrinking singular values of the weight matrix. In essence, by monitoring $MASED$, we can detect early signs of information degradation and modify training strategies to maintain the discriminative power of node embeddings.\\
The $MASED$ metric is defined over a graph $G$ with $N$ nodes as follows:
\begin{equation}\label{eq:mased_def}
MASED(G) = \frac{1}{N}\sum\limits_{i=1}^{i=N}{\frac{1}{N}\sum\limits_{j=1}^{j=N}}{\left(d^{euc}_{i,j}\right)^2} = \frac{1}{N^2}\sum\limits_{i=1}^{i=N}{\sum\limits_{j=1}^{j=N}}{\left(d^{euc}_{i,j}\right)^2},
\end{equation}
where $d^{euc}_{i,j}$ is the Euclidean distance between the representations/embeddings of node $i$ ($h^{(l)}_i$) and node $j$ ($h^{(l)}_j$).\\
Note that $MASED$ can be calculated over the output of each layer of the model under investigation. In order to simplify our analysis, we focus on the output of the $l$-th layer of a GCN, i.e., $ReLU(H^{(l)}) = ReLU(\hat{A}H^{(l-1)}W^{(l)})$.\\ 
In the rest of our analysis we calculate the Euclidean distance using the Gramian matrix ($G^{gram}$) of the embedding matrix $H^{(l)}$, i.e. $G^{gram} = H^{(l)} \cdot (H^{(l)})^T$. The Gramian takes that form because the rows of $H^{(l)}$ correspond to node embeddings. The value of the squared Euclidean distance between node embeddings (i.e., rows of matrix $H^{(l)}$) can be calculated using elements of $G^{gram}$, utilizing the following relationship:
\begin{equation}\label{eq:gram_eucl_rel}
    \left(d^{euc}_{i,j}\right)^2 = g_{i,i} -2g_{i,j} + g_{j,j},
\end{equation}
where $g_{i,j}$ is the $(i, j)$-element of $G^{gram}$ matrix.\\
For each node pair we calculate the squared Euclidean distance between its nodes using Equation \ref{eq:gram_eucl_rel}. Summing these values yields the total Euclidean squared distance as follows:
\begin{equation*}
    \sum\limits_{i=1}^N{\sum\limits_{j =1}^N{\left(d^{euc}_{i,j}\right)^2}} =  
    N \sum\limits_{i=1}^N{g_{i,i}} + \sum\limits_{i =1}^N{\sum\limits_{j =1}^N{-2 g_{i,j}}} + N \sum\limits_{j=1}^N{g_{j,j}} = 
\end{equation*}
\begin{equation*}
    N \cdot tr(G^{gram}) - 2 \cdot \mathbf{1}^T G^{gram} \mathbf{1} + N \cdot tr(G^{gram}) = 
\end{equation*}
\begin{equation}\label{eq:tot_eucl_sq_dist}
    2N \cdot tr(G^{gram}) - 2 \cdot \mathbf{1}^T G^{gram} \mathbf{1},
\end{equation}
where $\mathbf{1}$ is the all-one column vector. Equation \ref{eq:tot_eucl_sq_dist} holds because $g_{i,i}$ and $g_{j,j}$ do not depend on $j$ and $i$, respectively, and the double summation over $g_{i,j}$ is equivalent to the summation of every element of $G^{gram}$. Additionally, summing either of $g_{i,i}$ or $g_{j,j}$ yields the trace of $G^{gram}$.\\ 
Combining Equations \ref{eq:tot_eucl_sq_dist} and \ref{eq:mased_def} leads to the following expression of $MASED$:
\begin{equation}\label{eq:mased_initial}
    MASED^{(l)}(G) = \frac{2 \cdot tr(G^{gram})}{N} - \frac{2 \cdot \mathbf{1}^T G^{gram} \mathbf{1}}{N^2} = \frac{2}{N}\bigg(tr(G^{gram}) - \frac{1}{N}\mathbf1^T G^{gram}\mathbf1 \bigg).
\end{equation}
This alternative formulation of $MASED$ is particularly useful in deriving its upper and lower bounds, which are, in turn, important for controlling oversmoothing.

\subsection{Bounds on MASED}
In this subsection we derive the upper and lower bounds of $MASED$. To simplify the notation we drop the superscripts (i.e., we denote $H^{(l)}$ as $H$, $W^{(l)}$ as $W$ and $MASED^{(l)}(G)$ as $MASED(G)$), and introduce $\hat{H}^{(l-1)} = \hat{A} H^{(l-1)}$, the smoothed node embeddings after the step of averaging across neighboring nodes. These lead to the following expressions about $H^{(l)}$ and $G^{gram}$:
\begin{equation*}
    H^{(l)} = ReLU(\hat{A}H^{(l-1)}W^{(l)}) \equiv H = ReLU(\hat{H}W) \in \mathbb{R}^{N \times d}.
\end{equation*}
\begin{equation*}
    G^{gram} = HH^T \in \mathbb{R}^{N\times N},
\end{equation*}
whose diagonal elements can be written as $G^{gram}_{i,i} = ||H_{i,*}||_2^2 = r_i^2$, i.e., the diagonal elements of the Gramian are the squared norms of the node embeddings, where $H_{i,*}$ indicates the $i$-th row of $H$.\\
Considering the form of $G^{gram}$ we also derive the following:
\begin{equation*}
    \mathbf{1}^T G^{gram} \mathbf{1} = \mathbf{1}^T H H^T \mathbf{1} = \bigl( \mathbf{1}^T H\bigr) \bigl(\mathbf{1}^T H \bigr)^T = ||\mathbf{1}^T H||_2^2.
\end{equation*}
Hence, we can rewrite the terms of Equation \ref{eq:mased_initial} as:
\begin{itemize}
  \item \(tr(G^{gram})=\sum_{i=1}^N G^{gram}_{ii}=\sum_{i=1}^N r_i^2\).
  \item \(\tfrac1N\,\mathbf1^T G^{gram}\,\mathbf1
        =\sum_{j=1}^d\Bigl(\tfrac1{\sqrt N}\sum_{i=1}^N H_{i,j}\Bigr)^2
        =\sum_{j=1}^d z_j^2\).
\end{itemize}
Where
\begin{equation}\label{eq:r_z_definition}
r_i = \|H_{i,*}\|_2,
\quad
z_j = \frac{1}{\sqrt{N}}\sum_{i=1}^N H_{i,j},
\end{equation}
we recover the equivalent form of Equation \ref{eq:mased_initial}
\begin{equation}\label{eq:mased_final}
MASED(G) \;=\;\frac{2}{N}\Bigl(\sum_{i=1}^N r_i^2 \;-\;\sum_{j=1}^d z_j^2\Bigr).
\end{equation}

\subsubsection{Upper Bound}
In many cases, our objective is to drive $MASED$ as high as possible, since larger values of $MASED$ correspond to more diverse node embeddings, which, in turn, reduce oversmoothing.  However, it is equally important to understand the factors that limit $MASED$. An explicit upper bound acts as a guide to reveal which aspects of the model affect $MASED$ the most. \\
Considering the form of $MASED$ in Equation \ref{eq:mased_final} we observe that it is the result of the subtraction between positive quantities. Hence, the upper bound can be the upper bound of the first term of Equation \ref{eq:mased_final}. Using the Cauchy–Schwarz inequality and the upper bound of the norm of a product we conclude to the following Lemma.
\begin{lemma}\label{lem:upper_bound}
\begin{equation*}
    MASED(G) =\frac{2}{N}\Bigl(\sum_{i=1}^N r_i^2 \;-\;\sum_{j=1}^d z_j^2\Bigr) \leq \frac{2}{N} \sum_{i=1}^N r_i^2 \leq \frac{2}{N}N \max\limits_i r_i^2 \implies
\end{equation*}
\begin{equation*}
    MASED(G) \leq 2 \sigma^2_{\max}(W) \cdot M^2_{\hat{H}},
\end{equation*}
where $\sigma_{\max}(\cdot)$ denotes the largest singular value of the respective weight matrix, and $M_{\hat{H}} = \max\limits_i ||\hat{H}_{i,*}||_2$. In the above result we have used the fact that $r_i=\|\hat{H}_{i,*}W\|_2\le\sigma_{\max}(W)\|\hat{H}_{i,*}\|_2$, which holds for every row of $H$ (i.e., each node embedding).
\end{lemma}
By examining this bound, we observe how changes in each parameter, such the largest singular value of the weight matrix, or embedding norms, affects the maximum possible value of $MASED$. This analysis provides both a theoretical ceiling for $MASED$ and practical insight into which strategies are most effective for approaching that ceiling. Our result is aligned with the current line of research presented in Theorem \ref{theo:suzuki}. Additionally to that conclusion, Lemma \ref{lem:upper_bound} highlights the importance of the norms of the embeddings in increasing the upper bound of $MASED$. Hence, by increasing the largest singular value of the weight matrix and increasing embedding norms, we may allow the GNN to avoid oversmoothing. Conversely, when the largest singular values of the weight matrices or the norms of the node embeddings are small, $MASED$ is suppressed leading to smaller distances between node embeddings, and oversmoothing.

\subsubsection{Lower Bound}
While an upper bound on $MASED$ is important, especially when certain constraints or model parameters inherently limit how large $MASED$ can become, it does not provide a complete picture. In some cases, the upper bound may be loose or rarely attained in practice, offering limited insight into typical or guaranteed behavior.\\
In contrast, a meaningful lower bound is often more informative in contexts where the goal is to drive $MASED$ to higher values. A strong lower bound ensures that $MASED$ remains above a certain threshold under all valid conditions, providing a reliable performance guarantee. It reflects the worst-case scenario we can expect and helps us understand how much improvement is possible.\\
By focusing on the lower bound, we can identify which parameters or conditions most effectively raise this minimum and force the model to avoid oversmoothing. Note that trivially, $MASED$ is lower bounded by zero, when all node embeddings coincide, hence leading to zero distance between every pair of embeddings. We term that particular case as {\em extreme oversmoothing}, in which there is no variance in node embeddings and no classifier can properly distinguish them. Imposing the spectral-alignment condition, we conclude to the following Lemma.

\begin{lemma}\label{lem:lower_bound}
    The following lower bound on $MASED$ holds:
    \begin{equation*}
        MASED(G) \geq 2 \varepsilon \mathbb{E}[r^2] \geq 2 \varepsilon \cdot r_{\min}^2 \geq 2 \varepsilon \cdot \sigma^2_{\min}(W) \cdot m_{\hat{H}}^2,
    \end{equation*}
    where $r_{\min}$ denotes the smallest value of $r_i$'s, $\sigma_{\min}(W)$ denotes the smallest singular value of $W$, and $m_{\hat{H}} = \min\limits_i ||\hat{H}_{i,*}||_2$.
\end{lemma}
In order to prove Lemma \ref{lem:lower_bound}, we need to derive a more actionable lower bound of Equation \ref{eq:mased_final}. Hence, we define the following quantities:
\begin{align*}
      \mathbb{E}[r^2] &= \frac{1}{N}\sum_{i=1}^N r_i^2 = \frac{1}{N}tr(HH^\top) &\mathbb{E}[z^2] &= \frac{1}{d}\sum_{j=1}^d z_j^2
        = \frac{1}{dN}\mathbf1^\top HH^\top\mathbf1,
\end{align*}
where $\mathbb{E[\cdot]}$ denotes the expected value, and $r_i$ and $z_j$ are defined in Equation \ref{eq:r_z_definition}. Specifically, \(r_i\in\mathbb{R}^N\) measures the Euclidean norm of each row of \(H\), and \(z_j\in\mathbb{R}^d\) measures the normalized column‐wise sum.\\
As a result, Equation \ref{eq:r_z_definition} transforms to:
\begin{equation}\label{eq:mased_w_expectations}
    MASED(G) = 2 \bigg( \mathbb{E}[r^2] - \frac{d}{N} \mathbb{E}[z^2] \bigg).
\end{equation}
The above formula indicates the need of determining an upper bound of $\mathbb{E}[z^2]$, which, in turn, will lead to a lower bound of $MASED$.\\
For this, we exploit the spectral decomposition of the positive semi-definite matrix \(G^{gram} = H H^\top\), writing
\[
G^{gram} = \sum_{\ell=1}^N \lambda_\ell\,v_\ell v_\ell^\top,
\quad
\mathbf1^\top G^{gram}\,\mathbf1 = \mathbf1^\top HH^T\,\mathbf1 = \sum_{\ell=1}^N \lambda_\ell\,(v_\ell^\top \mathbf1)^2,
\]
where each $\lambda_\ell\ge0$ (i.e., eigenvalues of $G^{gram}$) and $\{v_\ell\}$ are orthonormal vectors (i.e., eigenvectors of $G^{gram}$). We isolate the contribution of the uniform direction \(\mathbf1\) relative to the total trace \(\sum_\ell \lambda_\ell\).  Here, \((v_\ell^\top\mathbf1)^2\) measures how closely each eigenvector aligns with the all‐ones pattern, and weighting by \(\lambda_\ell\) shows how strongly that direction influences the quadratic form \(\mathbf1^\top G^{gram}\,\mathbf1\), which, in turn, determines \(\mathbb{E}[z^2]\).
Intuitively, if a large fraction of the total “energy” (trace) of \(G^{gram}\) lies in the uniform direction, then the column averages \(z_j\) behave almost like constant multiples of the per‐row norms \(r_i\). In that case, the lower bound of $MASED$ is very close or equal to zero.\\
To avoid this case, we impose a spectral‐alignment condition. The spectral-alignment condition suggests that node embeddings have not collapsed to the same representation: there exists \(\varepsilon\in(0,1)\) such that
\[
\sum_{\ell=1}^N \lambda_\ell\,(v_\ell^\top \mathbf1)^2
\;\le\;(1-\varepsilon)\,N\,\sum_{\ell=1}^N \lambda_\ell.
\]
Note that since $G^{gram}$ is symmetric, the spectral theorem guarantees that its eigenvectors can be chosen to form an orthonormal basis. Hence, they are mutually orthogonal and each has unit norm (i.e., \(\|v_\ell\|_2=1\)), so \((v_\ell^\top \mathbf1)^2\le\|\mathbf1\|^2=N\).  The left-hand side is exactly the energy of the uniform direction with respect to \(G^{gram}\). If \(H_{i,*} = \beta_i\,u\) for some fixed \(u\), then \(v_1 = \mathbf1/\sqrt N\), \((v_1^\top\mathbf1)^2=N\), and 
\(\sum_\ell\lambda_\ell(v_\ell^\top\mathbf1)^2 = N\sum_\ell\lambda_\ell\). Our bound \(\le(1-\varepsilon) N \sum \lambda_\ell\) thus strictly excludes this degenerate case. The usage of, “\(\le\)” caps uniform alignment and guarantees genuine row‐diversity.\\
The spectral‐gap assumption is critical for preventing feature collapse in graph neural networks and for ensuring meaningful variability in node representations.  If every row of \(H\) were a scalar multiple of a single vector, then all node features would lie on a single line in \(\mathbb{R}^d\), and any aggregation or comparison based on dot‐products or learned linear mappings would render all representations indistinguishable up to scale.  By requiring a nonzero spectral gap (\(\varepsilon>0\)), we guarantee that at least an \(\varepsilon\)-fraction of the total row‐energy of \(H\) resides outside this degenerate direction, thereby preserving genuine discriminative structure.  Equivalently, since the all-ones vector \(\mathbf1\) encodes the uniform pattern “all rows identical”, bounding its squared projection onto the top eigenspace of \(HH^\top\) by \((1-\varepsilon)N\) ensures that no more than a \((1-\varepsilon)\)-fraction of the total variance in \(H\) can be explained by uniform alignment.\\
In turn, at least an \(\varepsilon\)-fraction of variance must lie in directions orthogonal to \(\mathbf1\), so that node features are not all “pointing the same way”.  This requirement is mild and generally satisfied in practical settings. Only pathological rank-one configurations (all rows exactly proportional, corresponding to \(\varepsilon=0\)) violate the gap. Therefore, a strictly positive spectral gap is a realistic and broadly applicable condition.\\
We therefore focus on \(\varepsilon>0\), noting that \(\varepsilon=0\) characterizes the pathological feature‐collapse scenario, where $MASED$ equals zero. The assumption of a positive $\varepsilon$ allows us to derive a positive lower bound, which highlights the contributing factors that can help increase the values of $MASED$. Since
\begin{equation*}
      \mathbb{E}[z^2]
      = \frac{1}{dN}\,\mathbf1^\top HH^\top\mathbf1 = \frac{1}{dN}\sum_{\ell=1}^N \lambda_\ell\,(v_\ell^\top \mathbf1)^2 \leq \frac{1-\varepsilon}{dN} \,N\,\sum_{\ell=1}^N \lambda_\ell \implies
\end{equation*}
\begin{equation*}
      \mathbb{E}[z^2] \leq \frac{1-\varepsilon}{dN} N\,tr(HH^\top)
      = (1-\varepsilon)\,\frac{N}{d}\,\mathbb{E}[r^2],
\end{equation*}
the alignment condition yields
\[
\mathbb{E}[z^2]
\;\le\;(1-\varepsilon)\,\frac{N}{d}\,\mathbb{E}[r^2] \implies 
\mathbb{E}[r^2]-\tfrac{d}{N}\,\mathbb{E}[z^2]
\;\ge\;\varepsilon\,\mathbb{E}[r^2].
\]
Substituting into the definition of $MASED$ in Equation \ref{eq:mased_w_expectations} gives
\begin{equation}\label{eq:mased_lower_bound}
    MASED(G)
    =2\Bigl(\mathbb{E}[r^2]-\tfrac{d}{N}\,\mathbb{E}[z^2]\Bigr)
    \;\ge\;2\varepsilon\,\mathbb{E}[r^2]
    \;\ge\; 0.
\end{equation}
This bound confirms that, under our spectral‐alignment assumption, the lower bound of $MASED$ is non-negative. Only in the degenerate case \(\varepsilon=0\), when the all‐ones vector is the top eigenvector and all embeddings collapse to one direction, does this bound collapse to the trivial lower bound, i.e., zero.

\noindent Using Equation \ref{eq:mased_lower_bound}, we, finally, derive the lower bound  presented in Lemma \ref{lem:lower_bound}. Lemma \ref{lem:lower_bound} shows the effect of the smallest singular value and the smallest norm of the node embeddings on the lower bound of $MASED$, which is related to oversmoothing appearance. Additionally, it highlights the tools available to reduce oversmoothing.

\subsection{Network‐Level Analysis of MASED}

Having established the upper (Lemma \ref{lem:upper_bound}) and lower (Lemma \ref{lem:lower_bound}) bounds on $MASED$ at each individual layer, we now extend these results to the entire network.  By tracking the evolution of $MASED$ from layer to layer, we derive global guarantees on how it changes from input to output. This extension is crucial as it reveals whether $MASED$ increases or decreases over multiple layers, and helps to identify the key layer-wise parameters that influence the network’s overall behavior.  In what follows, we show how the per-layer bounds combine and discuss the conditions under which the network preserves, amplifies, or shrinks the value of $MASED$.\\
In order to estimate the bounds of $MASED$ at the final layer of the model, we reintroduce the superscripts in our notation, i.e., we will bound $MASED^{(L)}(G)$ with $L$ being the model's depth. We also utilize Lemma \ref{lem:upper_bound} and Lemma \ref{lem:lower_bound} along with the inequality:
\begin{equation*}
    \sigma_{min}(B) ||u||_2 \leq ||Bu||_2 \leq \sigma_{max}(B) ||u||_2,
\end{equation*}
which holds for every matrix $B$ and vector $u$.\\

\noindent The lower bound (Lemma \ref{lem:lower_bound}) depends on  $m^{(L)}_{\hat{H}}$, which can be bounded as follows:
\begin{align*}
      m^{(L)}_{\hat{H}} = \min\limits_i ||\hat{H}^{(L)}_{i,*}||_2 = \min\limits_i ||\hat{A}H_{i,*}^{(L-1)}||_2 \geq \sigma_{min}(\hat{A}) \min\limits_i||H_{i,*}^{(L-1)}||_2 =\\ \sigma_{min}(\hat{A}) \min\limits_i||ReLU(\hat{A}H^{(L-2)}W^{(L-1)})||_2.
\end{align*}
Substituting $H^{(l)}$ telescopically leads to:
\begin{equation*}
    m^{(L)}_{\hat{H}} \geq \sigma^{L-1}_{min}(\hat{A}) \cdot \min\limits_i||X_{i,*}||_2 \cdot \prod\limits_{i=1}^{L-1} \sigma_{min}(W^{(i)}).
\end{equation*}
Similarly for $M^{(L)}_{\hat{H}}$ we get:
\begin{equation*}
    M^{(L)}_{\hat{H}} \leq \sigma^{L-1}_{max}(\hat{A}) \cdot \max\limits_i||X_{i,*}||_2 \cdot \prod\limits_{i=1}^{L-1} \sigma_{max}(W^{(i)}).
\end{equation*}
Combining the above results with Lemma \ref{lem:upper_bound} and Lemma \ref{lem:lower_bound} we arrive at the following conclusion about the bounds of $MASED^{(L)}$, i.e., the mean average squared distance of node representations at the final layer of the model:
\begin{equation}\label{eq:mased_output_bounds}
    \resizebox{\textwidth}{!}{
    $2 \varepsilon \bigg( \sigma^{L-1}_{min}(\hat{A}) \cdot m_{X} \cdot \prod\limits_{i=1}^{L} \sigma_{min}(W^{(i)}) \bigg)^2 \leq MASED^{(L)}(G) \leq 2\bigg(\sigma^{L-1}_{max}(\hat{A}) \cdot M_X \cdot \prod\limits_{i=1}^{L} \sigma_{max}(W^{(i)}) \bigg)^2$,
    }
\end{equation}
where $m_X = \min\limits_i||X_{i,*}||_2$, $M_X = \max\limits_i||X_{i,*}||_2$, and $X$ denotes the initial node features.\\

\noindent In Equation \ref{eq:mased_output_bounds}, the term \(\sigma_{\min}(\hat A)\), i.e., the smallest singular value of \(\hat A\) is often zero in real‐world graphs, leading the global lower bound to zero. For this reason, the per‐layer \(MASED^{(l)}(G)\) value is more useful.\\
Ensuring that each \(MASED^{(l)}(G)\) remains not only strictly positive but substantially above zero, particularly in the lower layers, helps preserve the variance of the input features and avoids the pitfalls of relying solely on the global \(MASED^{(L)}(G)\) lower bound.  While some reduction of \(MASED^{(l)}(G)\) in the upper layers may be expected or even beneficial for classification, maintaining substantially non-zero values in early layers is critical for robust feature propagation.  

\section{Implications of the Theoretical Analysis}
Equation \ref{eq:mased_output_bounds} highlights that $MASED$ is primarily influenced by two factors: the singular values of the weight matrices and their total number. To address the former, we propose a regularization method; namely \emph{G-Reg}, as a means of increasing the singular values, while for the latter we provide further intuition and suggest reducing the number of weight matrices as a practical remedy.

\subsection{\emph{G-Reg}: Regularization of the Standard Deviation of the Weight Matrix}

An idea for reducing oversmoothing, born from the results of our theoretical analysis, is to introduce a novel regularization approach tailored to graph‐based models. Existing regularizers (e.g., standard weight decay or dropout) do not explicitly address the reduction of the singular values of the weight matrices, which have been shown to play an important role in oversmoothing. If the rows of \(W^{(l)}\) tend to become linearly dependent, the smallest singular value $\sigma_{\min}\bigl(W^{(l)}\bigr)$ will monotonically decrease toward zero, which, in turn, collapses the lower bound of Equation \ref{eq:mased_output_bounds} to zero.\\
To address this issue, we propose \emph{G-Reg}, which aims to reduce the linear dependence of the rows of the weight matrices by rewarding large standard deviation among the elements of each row. Increasing the standard deviation can thus increase the directional diversity of the matrix rows. This results is achieved under the assumption of row-wise independent perturbations, which means that each row is randomly perturbed independently of the others. Since linear independence of the rows implies that the matrix has full (or almost full) row rank, it follows that the smallest singular value $\sigma_{\min}(W^{(l)}) > 0$. Positive $\sigma_{\min}(W^{(l)})$ values allow the lower bound of Equation \ref{eq:mased_output_bounds} to remain positive as well, pushing the average node distances to higher values.\\
\noindent Formally, let \(W^{(l)}\) denote the learnable weight matrix at layer \(l\), we define the \emph{G-Reg} penalty as
\begin{equation}
    \mathcal{L}_{\mathrm{G\text{-}Reg}}=
    \lambda_w \sum_{l=1}^{L}\frac{1}{d}\sum\limits_{i=1}^d{std(W^{(l)}_{i,*})},
\end{equation}
where $std$ denotes the standard deviation, $W^{(l)}_{i,*}$ is the $i$-th row of matrix $W^{(l)}$, and \(\lambda_w>0\) is a tunable strength parameter.

\subsection{Effect of multiple weight matrices on MASED}\label{sec:4.4}
Given the importance of weight values, as mentioned above, we study further the effect of the number of weight matrices through which the input signal passes. Each weight matrix can be regarded as a transformation from one embedding space to another. Hence, the number of weight matrices expresses the number of embedding spaces through which the input signal passes.\\
Increasing the depth of GCNs introduces a significant risk of losing critical information even before training begins. This phenomenon arises from the probabilistic properties of random weight matrices at initialization. Typically, these matrices are initialized with values drawn from a Gaussian distribution, resulting in eigenvalues within the unit circle and singular values centered around 2.\\
Suppose that the top-\(k\) singular values of each weight matrix exceed a threshold (e.g., 0.5). Then the corresponding input feature dimensions are only mildly weakened as they pass through each layer. Conversely, feature dimensions tied to smaller singular values shrink quickly in deep networks. As the network depth \(L\) increases, the probability that a useful feature direction retains sufficient strength across all layers diminishes exponentially, limiting its influence on the final output. Specifically, if the weight matrices have a width $d > k$ ($k$ once again denoting the number of singular values that exceed a threshold), the probability $Pr[Q]$ that an informative direction survives through all layers is given by:
\begin{equation*}
    Pr[Q] = \left( \frac{k}{d} \right)^L = \beta^L, \quad \beta < 1.
\end{equation*}
This exponential decay implies that deeper networks are increasingly likely to suppress informative features before any learning occurs. Consequently, the model may be ``doomed to fail", as essential information is lost during the initial forward passes, leading to suboptimal performance that cannot be improved through subsequent training. The underlying issue is that the initial random weight matrices, combined with the depth of the network, effectively filter out significant components of the input features. As a result, gradient signals weaken or vanish entirely, preventing the model from updating its parameters effectively. Consequently, increasing the number of weight matrices can degrade overall performance.\\
This is also related to the value of $MASED$, as shown in Equation \ref{eq:mased_output_bounds}. As the number of weight matrices increases, the number of factors in the products of the smallest and largest singular values also increases. To keep $MASED$ large, each weight matrix should maintain both a large minimum and maximum singular value. This becomes increasingly difficult as more layers are added. Therefore, while using multiple weight matrices allows the model to capture complex input relationships, too many weight matrices might lead to performance deterioration.

\subsection{Decoupling Weight Matrices from Adjacency Powers}
\noindent Equation \ref{eq:mased_output_bounds} additionally shows that the distance between node embeddings after graph convolution depends on both the number of weight matrices and the adjacency power used for aggregation. Conventionally, each layer \(l\) has a different weight matrix \(W^{(l)}\), coupling receptive field size to parameter count. Yet, successive adjacency multiplications already capture high-order structure, making the use of a separate \(W^{(l)}\) for each layer not always beneficial. This was discussed in Subsection \ref{sec:4.4}.\\
Based on this observation, we explore the possibility of aggregating an extended  multi-hop neighborhood via a single weight matrix, thereby reducing the number of matrices in the network. This principle resembles the APPNP \citep{APPNP} single-matrix propagation scheme. According to our analysis, reducing the number of matrices can also limit the extent of oversmoothing.
\begin{figure}[H]
    \centering
    \includegraphics[width=.797\textwidth]{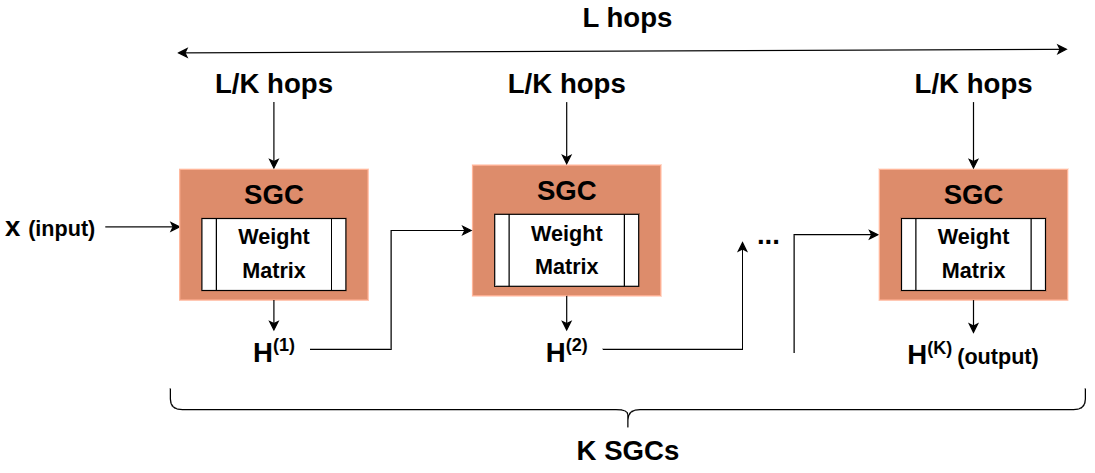}
    \caption{Each SGC layer uses the adjacency matrix raised to the power of $L/K$, and takes as input either the output of the previous layer or the initial node features if it is the first layer.}
    \label{fig:reduced_weight_matrices}
\end{figure}
\noindent In practice, we distribute the number of hops $L$ that we want to capture, i.e., the distance from a single node, across $K$ stacked blocks of SGC, each aggregating up to $L/K$ hops before applying its own transformation. A visual representation of this mechanism is provided in Figure \ref{fig:reduced_weight_matrices}. For example, if \(L = 10\) (i.e., we wish to aggregate information up to 10 hops away), using two stacked layers of SGC implies that each layer aggregates information up to \(L/2 = 5\) hops away. This reduces parameter redundancy and enables multi-hop feature mixing.

\section{Experiments}
In this section we perform a series of experiments, inspired by the theoretical results presented above. The experiments aimed to confirm the power of $MASED$ in quantifying oversmoothing and the reduction of the problem by the proposed changes in the learning process.
\subsection{Experimental Setup}
\textbf{Datasets:} Aligned to most of the literature, we focused on seven well-known benchmarks: \textit{Cora, CiteSeer, Pubmed, Photo, Computers, Physics} and \textit{CS}. For the first three co-citation datasets we used the same data splits as in \citet{Kipf_gcn}, where all the nodes except the ones used for training and validation are used for testing. For the \textit{Photo, Computers, Physics} and \textit{CS} datasets we followed the same splits as in \citet{photo_dataset}. Dataset statistics can be found in Appendix \ref{apdx:G}. \\
\textbf{Models:} We experimented with the architectures of GCN \citep{Kipf_gcn}, ResGCN \citep{resgcn}, and SGC \citep{sgc}.\\ 
\textbf{Hyperparameters:} We performed a hyperparameter sweep (details in Appendix \ref{apdx:H}) to determine the optimal hyperparameter values, based on their performance on the validation set. For GCN and ResGCN, we set the number of hidden units for each layer to 128 across all benchmark datasets. For SGC when using a single layer, the input dimension equals the number of features while the output dimension is the number of classes. When using multiple layers of SGC the number of hidden units is also 128. $L_2$ regularization was applied with a penalty of $5\cdot10^{-4}$, and the learning rate was set to $10^{-3}$ for GCN and ResGCN, while for SGC the optimal value was $6 \cdot 10^{-3}$. Depth varied between 2 and 40 layers depending on the setting under investigation. Finally, $\lambda_w \in \{ 0, 2, 3, 4, 6, 8 \}$ for GCN and ResGCN, while for SGC $\lambda_w \in \{ 0, 0.01, 0.5, 1, 2 \}$.\\
\textbf{Configuration:} Each experiment was run 10 times and we report the average performance over these runs. We trained all models within 200 epochs using Cross Entropy as a loss function.

\subsection{Experimental Results}

\begin{figure}[h]
    \centering
    \includegraphics[width=\textwidth]{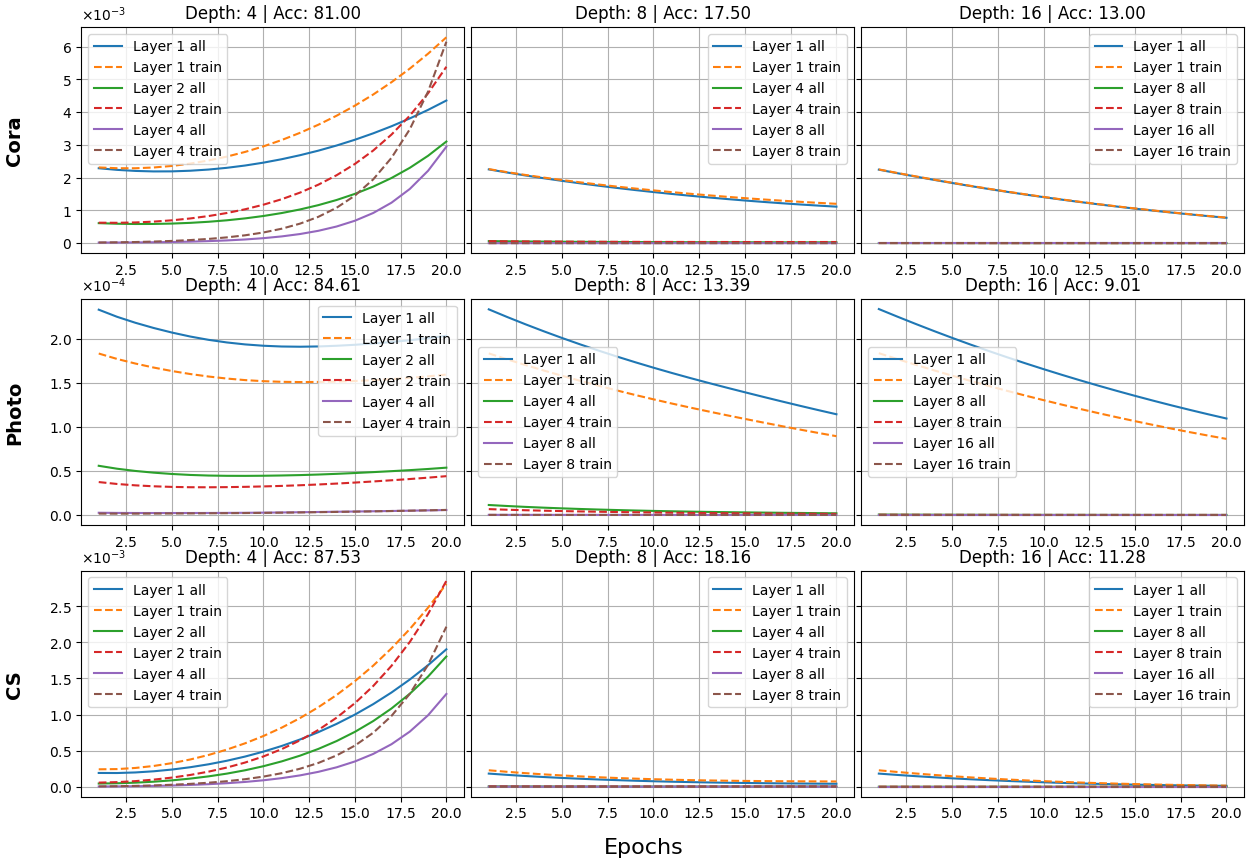}
    \caption{Epoch evolution of the Mean Average Squared Euclidean Distance ($MASED$) value of the embeddings of all nodes and training nodes separately. We show results for 3 different depths of a GCN model, illustrating how $MASED$ changes in the first, the middle and the last layer of the model. We also include the accuracy achieved by each model.}
    \label{fig:MASED_GCN}
\end{figure}

\noindent \textbf{MASED evolution at different network depths:}\\
Based on Equation~ \ref{eq:mased_output_bounds}, we investigate the extent to which $MASED$ exhibits the predicted scaling, with larger values in early layers and smaller ones in deeper layers. Figure \ref{fig:MASED_GCN} presents the evolution of $MASED$ across training epochs for a plain GCN at depths 4, 8, and 16. Separate curves are shown for training nodes and for all nodes for each layer on \textit{Cora, Photo}, and \textit{CS} datasets. At depth 4, $MASED$ increases over the 20 epochs, driven most strongly by the first layer while at deeper layers it rises more slowly. This upward trend indicates that feature values are diverging before being mixed in later layers. At depths 8 and 16, all layers show a steady decrease in $MASED$. In subsequent experiments we investigate whether this behavior aligns with embedding norms as Equation~ \ref{eq:mased_output_bounds} suggests.\\ 
These findings show that $MASED$ is most informative at the first layer, its rise or fall signaling whether the network can expand or shrink embedding differences. Similar plots are presented for the rest of the datasets along with similar plots for ResGCN and SGC in Appendix \ref{apdx:C}. Note that Figure \ref{fig:MASED_GCN} shows only the first 20 epochs; the complete results over 200 epochs can be found in Appendix \ref{apdx:C}.\\
Furthermore, these observations align with the mathematical bounds derived from Equation~ \ref{eq:mased_output_bounds}: both the upper and lower bounds scale as $\beta^L$, where $L$ is the layer index and $\beta$ is an expression depending on weight matrix properties and the underlying graph topology. Hence the first layer is generally allowed to attain larger values, while deeper layers are more prone to smaller values.\\

\begin{figure}[h]
    \centering
    \includegraphics[width=\textwidth]{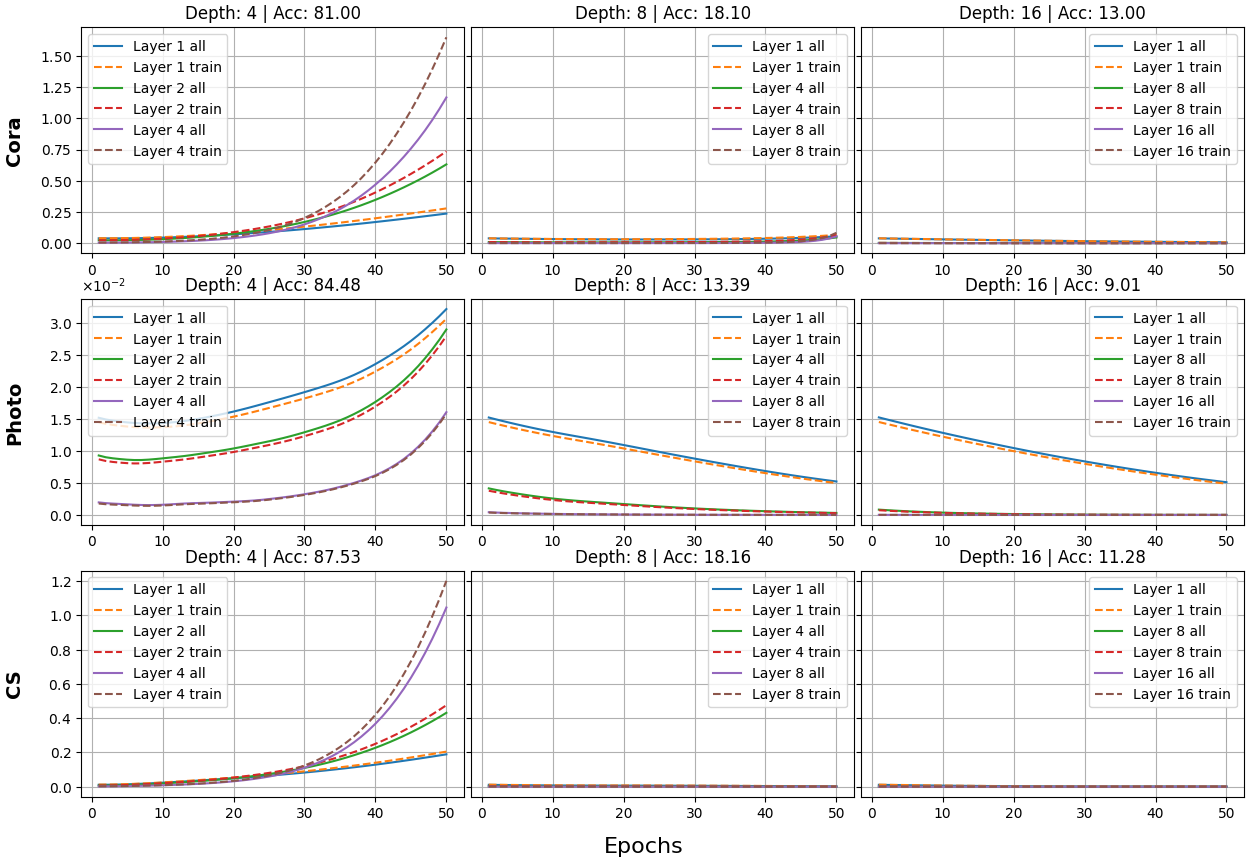}
    \caption{Epoch evolution of the average value of the norms of the embeddings of all nodes and of the training nodes separately. We show results for 3 different depths of a GCN model and average norm values in different layers within the model. We show how norms evolve in the first, the middle and the last layer of each model. We also include the accuracy achieved by each model.}
    \label{fig:norms_GCN}
\end{figure}

\begin{figure}[h]
    \centering
    \includegraphics[width=\textwidth]{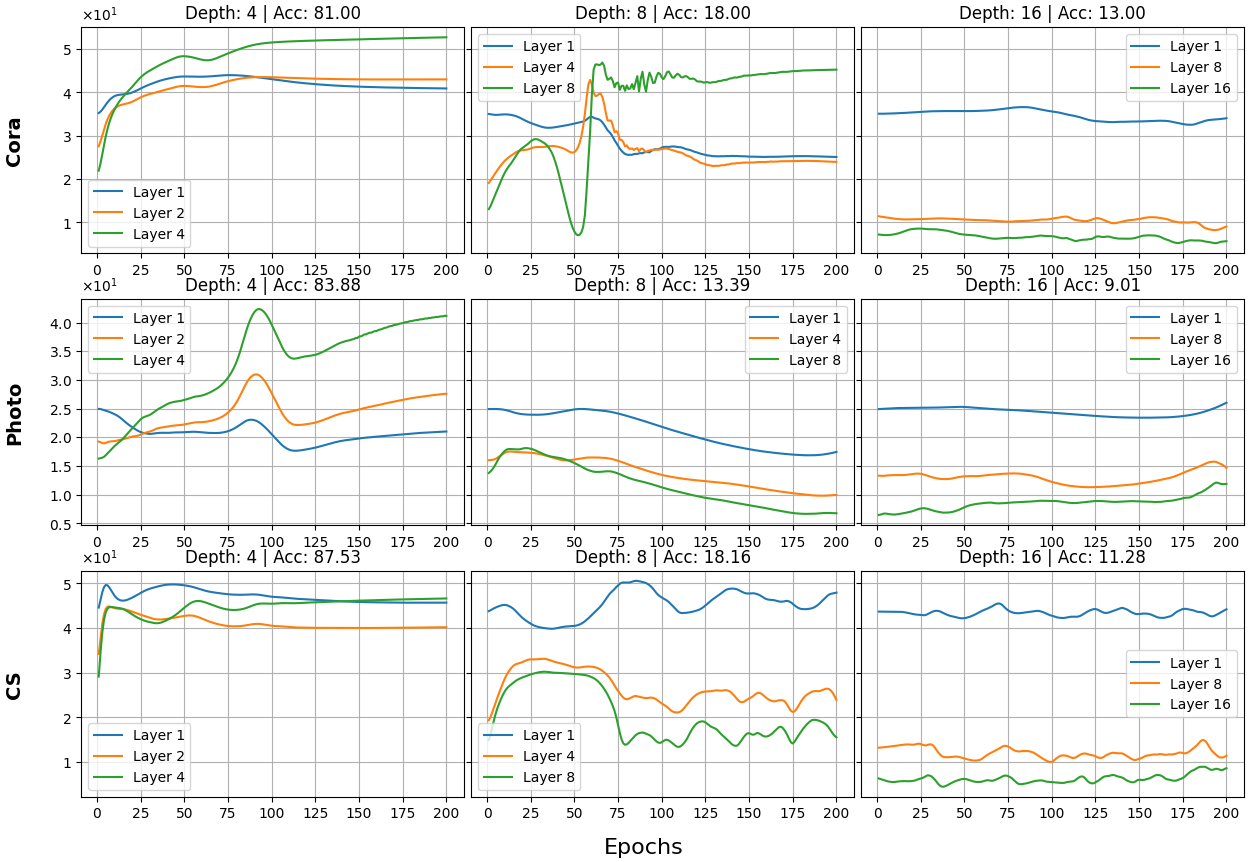}
    \caption{Epoch evolution of the average value of the angles between the class centroids of the embeddings of the training nodes. We show results for 3 different depths of a GCN model and average norm values in different layers within the model. We show how angles evolve in the first, the middle and the last layer of each model. We also include the accuracy achieved by each model.}
    \label{fig:angles_GCN}
\end{figure}

\noindent \textbf{Norms and angles of embeddings at different network depths:}\\ Building on the previous experimental results for $MASED$, we now investigate how norms influence its values and examine whether this behavior aligns with the proposed theoretical predictions. Figure \ref{fig:norms_GCN} plots the average norm of node embeddings for a GCN at depths 4, 8 and 16 on the \textit{Cora, Photo} and \textit{CS} datasets. At depth 4 (high accuracy), the average norm of node embeddings steadily rises over epochs, reflecting growing feature norms and healthy propagation at moderate depth; at depth 8 (smaller accuracy), the increase is far more subdued, its curve remaining nearly flat and indicating early onset of oversmoothing that limits further norm growth; at depth 16 (smallest accuracy), the average norm of node embeddings is essentially constant, signifying collapse of all node representations and complete oversmoothing. The gap between the trajectories for shallow and deep models underscores the sensitivity of norm dynamics to depth, with deeper networks rapidly losing the capacity to amplify node signals effectively. Similar plots for the rest of the datasets, along with the residual GCN (ResGCN) and SGC variants are shown in Appendix \ref{apdx:B}. Note that Figure \ref{fig:norms_GCN} shows only the first 50 epochs; the complete results over 200 epochs can be found in Appendix \ref{apdx:B}.\\
Figure \ref{fig:angles_GCN} presents the average angle between the class centroids of the training nodes, i.e., the centroids of each class of the training nodes on \textit{Cora, Photo} and \textit{CS} datasets. At depth 4 fluctuations start at moderate amplitude and then almost disappear, indicating stable gradient flow and smooth convergence as feature norms increase; at depth 8 the curves have larger early spikes and a slower decay, reflecting instability from deeper aggregation and a more unstable training process; at depth 16 fluctuations drop to near zero almost immediately, mirroring the flat average norm of node embeddings and showing that extreme oversmoothing not only suppresses norm growth but also prevents meaningful parameter updates. These small fluctuations at larger depths, combined with the flat (and almost zero) average norm highlight the key role of the embedding norms in reducing oversmoothing (in agreement with Equation \ref{eq:mased_output_bounds}). One would expect that if the angles between embeddings remain non-zero then the model would be capable of solving the underlying task. However, we observe that if the norms of the embeddings become very small, then the input signal information is lost and the angles between node embeddings do not suffice to capture the differences between node classes. Similar plots for the rest of the datasets and for ResGCN and SGC are provided in Appendix \ref{apdx:B}.\\

\begin{figure}[h]
    \centering
    \includegraphics[width=\linewidth]{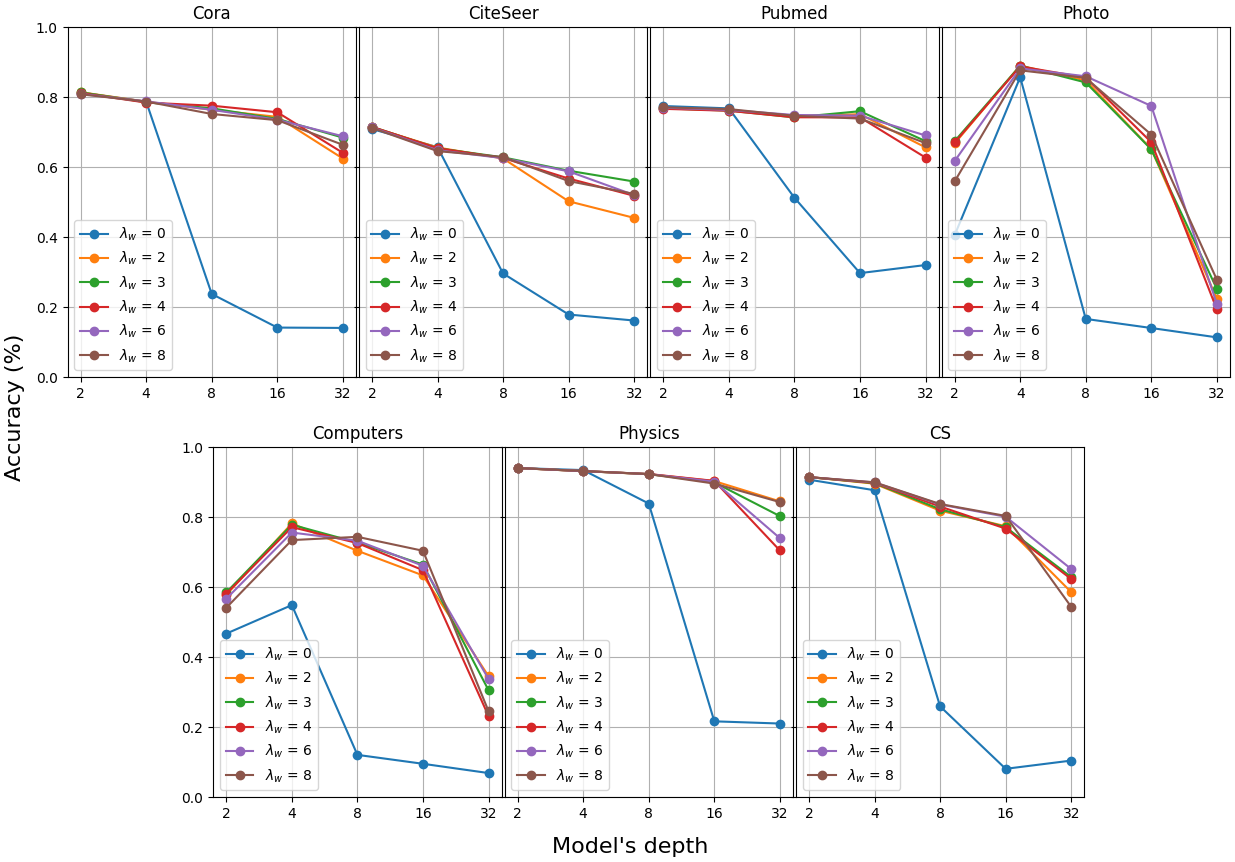}
    \caption{GCN with and without the proposed \emph{G-Reg} regularization across 7 datasets for varying depth. We include results for different values of $\lambda_w$.}
    \label{fig:benchmarks_GCN}
\end{figure}

\noindent \textbf{Reducing oversmoothing through regularization:}\\  
Figure \ref{fig:benchmarks_GCN} shows node classification accuracy of a GCN on each dataset. Model depth varies on the horizontal axis and regularization strength \(\lambda_{w}\) of the proposed \emph{G-Reg} is encoded by curve color. In all seven subplots the unregularized baseline (\(\lambda_{w}=0\)) peaks at shallow depths and then declines sharply. On the contrary, GCNs with \(\lambda_{w}>0\) resist oversmoothing and achieve much higher accuracy than their unregularized counterpart. These results confirm that by rewarding larger standard deviation of the weight rows, through \emph{G-Reg}, oversmoothing can be reduced, permitting effective propagation at depths where the unregularized models fail. In particular, the proposed method enables deep architectures that resist oversmoothing and remain capable of solving node classification tasks at large depths. Similar plots for ResGCN and SGC appear in Appendix \ref{apdx:D}.\\
Our empirical observations align with the theoretical bounds derived from Equation \ref{eq:mased_output_bounds}, which predict that as depth \(L\) increases, both the upper and lower bounds on $MASED$ shrink, thereby inducing oversmoothing. The proposed regularization reduces the co-linearity of weight matrix rows, which, in turn, increases the smallest singular value. As a consequence, the lower bound will increase, leading to larger $MASED$ values and variance of node embeddings in deeper layers. In this way, the regularization counteracts the depth‐induced tightening of representational limits and enables the model to reduce oversmoothing.\\

\begin{table}[h]
    \centering
    \caption{Comparison of different methods with and without the proposed regularization in the “cold start” scenario. Only the features of the nodes in the training set are available to the model. We present the best accuracy (i.e., Acc.) of the model and the depth (i.e., \# Layers) at which this accuracy was achieved, for GCN, ResGCN and SGC.}
    \resizebox{\textwidth}{!}{\begin{tabular}{|c|c c c|c c c|c c c|}
        \hline
        \multirow{3}{*}{Dataset} & \multicolumn{3}{c|}{\textbf{\underline{GCN}}} & \multicolumn{3}{c|}{\textbf{\underline{ResGCN}}} & \multicolumn{3}{c|}{\textbf{\underline{SGC}}}\\
        {} & {} & {} & {} & {} & {} & {} & {} & {} & {}\\
        {} & $\lambda_{w}$ & Acc.(\%) \& std & \#L & $\lambda_{w}$ & Acc.(\%) \& std & \#L & $\lambda_{w}$ & Acc.(\%) \& std & \#L\\
        \hline
        \multirow{2}{*}{Cora} & 
        0 & 60.50 \scriptsize$\pm$ 4.4 & 4 & 0 & 69.13 \scriptsize$\pm$ 0.9 & 6 & 0 & 61.16 \scriptsize$\pm$ 0.4 & 5\\
        {} & 8 & \textbf{73.26 \scriptsize$\pm$ 0.9} & 19 & 3 & \textbf{73.88 \scriptsize$\pm$ 0.8} & 29 & 1 & \textbf{65.68 \scriptsize$\pm$ 1.6} & 8\\
        \hline
        \multirow{2}{*}{CiteSeer} &
        0 & 41.95 \scriptsize$\pm$ 0.2 & 4 & 0 & 45.85 \scriptsize$\pm$ 1.2 & 7 & 0 & 38.86 \scriptsize$\pm$ 0.1 & 7\\
        {} & 4 & \textbf{48.08 \scriptsize$\pm$ 1.3} & 25 & 2 & \textbf{47.97 \scriptsize$\pm$ 1.2} & 23 & 1 & \textbf{49.79 \scriptsize$\pm$ 0.1} & 17\\
        \hline
        \multirow{2}{*}{Pubmed} &
        0 & 60.81 \scriptsize$\pm$ 3.9 & 4 & 0 & 69.23 \scriptsize$\pm$ 0.5 & 6 & 0 & 63.57 \scriptsize$\pm$ 0.1 & 6\\
        {} & 4 & \textbf{72.15 \scriptsize$\pm$ 0.7} & 25 & 2 & \textbf{71.22 \scriptsize$\pm$ 1.1} & 23 & 0.01 & \textbf{64.51 \scriptsize$\pm$ 0.1} & 6\\
        \hline
        \multirow{2}{*}{Physics} &
        0 & 51.12 \scriptsize$\pm$ 7.9 & 3 & 0 & 82.45 \scriptsize$\pm$ 0.6 & 6 & 0 & \textbf{74.54 \scriptsize$\pm$ 1.3} & 5\\
        {} & 8 & \textbf{89.98 \scriptsize$\pm$ 0.7} & 23 & 8 & \textbf{89.98 \scriptsize$\pm$ 0.7} & 32 & 0.01 & 74.17 \scriptsize$\pm$ 3.6 & 5\\
        \hline
        \multirow{2}{*}{CS} &
       0 & 14.11 \scriptsize$\pm$ 8.2 & 4 & 0 & 47.63 \scriptsize$\pm$ 9.8 & 6 & 0 & 71.43 \scriptsize$\pm$ 1.5 & 7\\
        {} & 8 & \textbf{77.48 \scriptsize$\pm$ 2.4} & 18 & 8 & \textbf{79.51 \scriptsize$\pm$ 1.4} & 19 & 0.5 & \textbf{73.80 \scriptsize$\pm$ 0.1} & 7\\
        \hline
        \multirow{2}{*}{Photo} &
        0 & 20.99 \scriptsize$\pm$ 9.8 & 2  & 0 & 27.14 \scriptsize$\pm$ 7.9 & 17 & 0 & 45.06 \scriptsize$\pm$ 4.3 & 2\\
        {} & 2 & \textbf{81.33 \scriptsize$\pm$ 2.8} & 7 & 2 & \textbf{84.16 \scriptsize$\pm$ 0.9} & 6 & 0.5 & \textbf{48.91 \scriptsize$\pm$ 6.6} & 2\\
        \hline
        \multirow{2}{*}{Computers} &
        0 & 18.85 \scriptsize$\pm$ 9.9 & 18 & 0 & 15.84 \scriptsize$\pm$ 9.4 & 2 & 0 & 7.97 \scriptsize$\pm$ 3.5 & 2\\
        {} & 4 & \textbf{69.74 \scriptsize$\pm$ 6.3} & 10 & 3 & \textbf{71.92 \scriptsize$\pm$ 3.0} & 8 & 0.5 & \textbf{9.6 \scriptsize$\pm$ 2.8} & 2\\
        \hline
    \end{tabular}}
    \label{tab:cold_start}
\end{table}

\noindent \textbf{Performance under the ``cold start” scenario:}\\
Table \ref{tab:cold_start} reports the best accuracy achieved by each of the three models (GCN, ResGCN, and SGC) on each dataset, under the ``cold start" setup, where only the labeled nodes have features initially. The results are presented together with the corresponding \(\lambda_{w}\) value, and the depth \(\#L\) at which each model attains that performance. We observe that nonzero \(\lambda_{w}\) consistently achieves better performance in larger depths. In particular, regularized models often attain peak performance at depths two to five times greater than the unregularized baselines and consistently increase accuracy by a statistically significant amount. This pattern is consistent across all three models, including the ones that are considered tolerant towards oversmoothing. In cold‐start experiments, where unlabeled node features are zeroed out, unregularized models are restricted to very shallow architectures, whereas models with optimized \(\lambda_{w}\) achieve their best results at much deeper configurations. These findings demonstrate that the proposed regularization not only improves overall accuracy but also enables deeper GNNs and effectively leverages additional propagation steps under both standard and feature‐scarce conditions.\\

\begin{figure}[h]
    \centering
    \includegraphics[width=\linewidth]{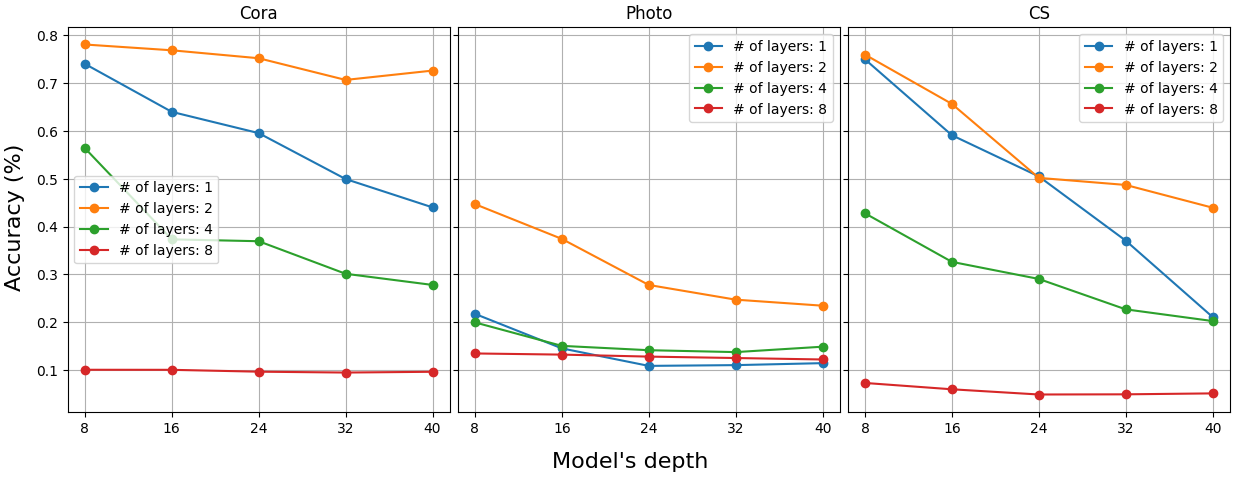}
    \caption{Comparison of SGC models with varying number of stacked SGC layers, across 3 different datasets for varying depth. At every depth all models have access to the same information. We only vary the number of trainable weight matrices (i.e., the number of SGC layers).}
    \label{fig:varying_SGCs}
\end{figure}
 
\noindent \textbf{Varying the number of SGC layers at different network depths:}\\  
Figure \ref{fig:varying_SGCs} compares SGC accuracy on \textit{Cora, Photo} and \textit{CS} datasets as the number of trainable weight matrices (SGC layers), but also the overall depth of the model (i.e., number of hops of each node's neighborhood) change. Across all datasets, the 2‐layer configuration consistently delivers the highest accuracy, followed by the 1-layer, then the 4-layer, and finally the 8-layer, which performs the worst. This ordering reflects the need for sufficient trainable layers (i.e., more than one) to perform the necessary feature transformations, while avoiding too many trainable weight matrices which can have negative effect. In addition to achieving the highest accuracy, the 2-layer configuration exhibits higher resistance to oversmoothing, as the depth of the model increases, as compared to other models, e.g. the 1-layer one. Results for the remaining datasets appear in Appendix \ref{apdx:E}.\\
These results reinforce the argument of subsection \ref{sec:4.4} that too many layers can harm performance by introducing redundant weight matrices. In our experiments, two layers strike the optimal balance, demonstrating that flexible assignment of the total number of hops to a small number of weight matrices is essential for deep graph models.\\

\section{Conclusion}
We have proposed the use of Mean Average Squared Euclidean Distance ($MASED$) between node embeddings, as a way to quantify the extend of oversmoothing in GNNs. We further derived layer‐wise bounds on $MASED$ and shown how they combine across depth to derive the global upper and lower bounds. Based on those bounds, we have highlighted the importance of the norms of the node embeddings and the key role of both the largest and the smallest singular values of the weight matrices. A nonzero smallest singular value can prevent feature collapse and ensures a meaningful lower bound on $MASED$, which, in turn, preserves variance among node representations and gradient flow.\\
Furthermore, we have shown that tying the number of trainable weight matrices directly to the total number of hops causes redundancy and oversmoothing in deep GNNs. Our theoretical bounds from Equation \ref{eq:mased_output_bounds} explain this effect and motivate reducing the number of trainable weight matrices to a number that is much lower than the total number of hops. We have also introduced \emph{G-Reg}, a regularization method, which penalizes the small standard deviation between the rows of the weight matrices, hence leading to larger smallest singular values, which, in turn, increase the bounds of Equation \ref{eq:mased_output_bounds}. We have conducted an extensive set of experiments which showed that these strategies improve accuracy and robustness, even when combined with methods that resist oversmoothing in different ways.\\
The theoretical analysis presented in this paper opens up a multitude of possible research options to address the problem of oversmoothing. One such direction that we consider important is the interaction between different existing approaches against oversmoothing. $MASED$ highlights the influence of weight matrix singular values and norms, providing a principled way to quantify the problem. Leveraging $MASED$ as a common evaluation tool will enable a systematic exploration of how architectural changes, normalization techniques, and activation adjustments interact, and whether they can be combined in a complementary manner to enable deeper GNNs.


\acks{The research work was supported by the Hellenic Foundation for Research and Innovation (HFRI) under the 4th Call for HFRI PhD
Fellowships (Fellowship Number: 10860).}


\newpage

\appendix
\section{Lower Bound on \(\Delta_r\)}\label{apdx:A}

Let
\[
r_i = \|H_{i,*}\|_2 = \|\hat{H}_{i,*}W\|_2,\quad
M_{\hat{H}} = \max_{1\le i\le N}\|\hat{H}_{i,*}\|_2,\quad
m_{\hat{H}} = \min_{1\le i\le N}\|\hat{H}_{i,*}\|_2,
\]
and define the row–norm spread
\(\displaystyle \Delta_r = r_{\max}-r_{\min}\).

\subsubsection*{Step 1: Bounds on \(r_{\max}\) and \(r_{\min}\)}

From the singular‐value inequalities,
\[
\|\hat{H}_{i,*}W\|_2 \;\ge\;\sigma_{\min}(W)\,\|\hat{H}_{i,*}\|_2,
\quad
\|\hat{H}_{i,*}W\|_2 \;\le\;\sigma_{\max}(W)\,\|\hat{H}_{i,*}\|_2.
\]
Choose \(i^*\) such that \(\|\hat{H}_{i^*,*}\|_2=M_{\hat{H}}\).  Then
\[
r_{\max} = \max_i r_i \;\ge\; r_{i^*}
= \|\hat{H}_{i^*,*}W\|_2
\;\ge\;\sigma_{\min}(W)\,M_{\hat{H}}.
\]
Similarly, pick \(j^*\) with \(\|\hat{H}_{j^*,*}\|_2=m_{\hat{H}}\).  Then
\[
r_{j^*} = \|\hat{H}_{j^*,*}W\|_2 \;\le\;\sigma_{\max}(W)\,m_{\hat{H}},
\]
and since \(r_{\min} = \min_i r_i \le r_{j^*}\), we have
\[
r_{\min} \;\le\;\sigma_{\max}(W)\,m_{\hat{H}}.
\]

\subsubsection*{Step 2: Subtract to obtain \(\Delta_r\)}

Subtracting the two bounds (lower bound on \(r_{\max}\) minus upper bound on \(r_{\min}\)) gives
\[
\Delta_r
= r_{\max} - r_{\min}
\;\ge\;
\sigma_{\min}(W)\,M_{\hat{H}}
\;-\;\sigma_{\max}(W)\,m_{\hat{H}}.
\]

\subsubsection*{Step 3: Express via the condition number}

Define the condition number \(\kappa(W)=\sigma_{\max}(W)/\sigma_{\min}(W)\).  Then
\(\sigma_{\max}(W)=\kappa(W)\,\sigma_{\min}(W)\), and the bound becomes
\[
\Delta_r \;\ge\;\sigma_{\min}(W)\,\bigl(M_{\hat{H}} - \kappa(W)\,m_{\hat{H}} \bigr).
\]
This is nonnegative (and thus meaningful) whenever \(M_{\hat{H}}\ge\kappa(W)\,m_{\hat{H}}\),  
i.e.\ when the inherent spread in \(\hat{H}\) exceeds the distortion introduced by \(W\). We make that assumption in order to further proceed our analysis.

\section{MASED Evolution plots}\label{apdx:C}

\begin{figure}[H]
    \centering
    \includegraphics[width=\linewidth]{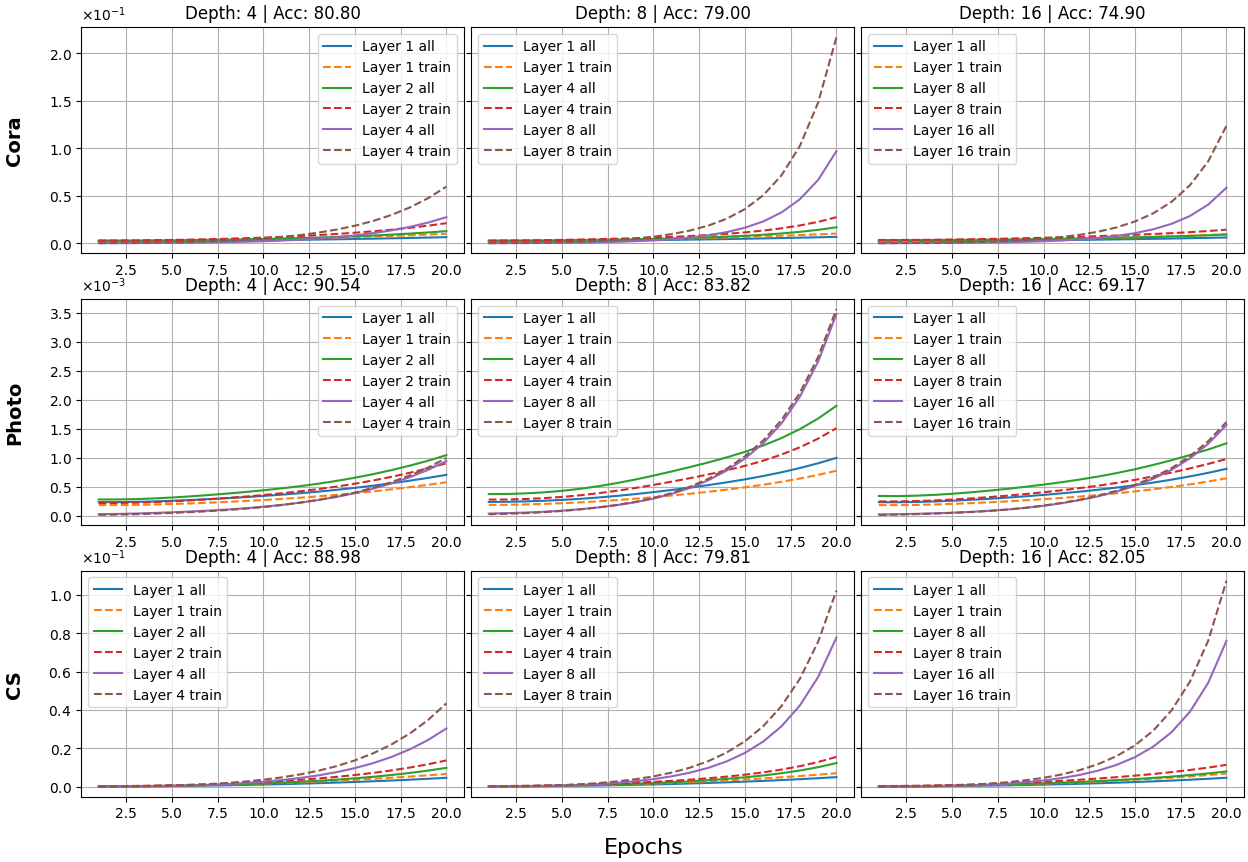}
    \caption{Epoch evolution of the Mean Average Squared Euclidean Distance ($MASED$) value of the embeddings of all nodes and training nodes. We show results for 3 different depths of a ResGCN model and $MASED$ values in different layers within the model. We show how $MASED$ evolve in the first, the middle and the last layer of each model. We also include the accuracy achieved by each model.}
    \label{fig:mads_resgcn}
\end{figure}

\begin{figure}[H]
    \centering
    \includegraphics[width=\linewidth]{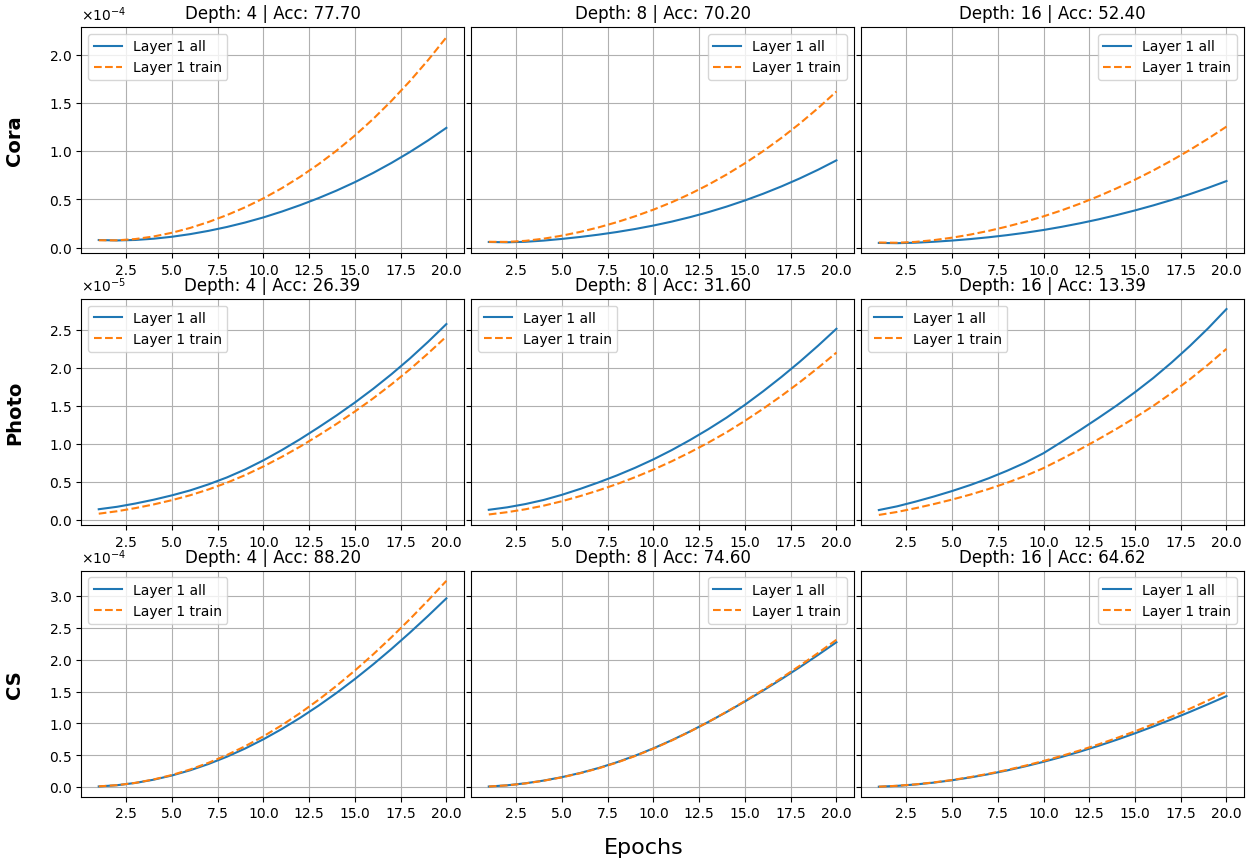}
    \caption{Epoch evolution of the Mean Average Squared Euclidean Distance ($MASED$) value of the embeddings of all nodes and training nodes. We show results for 3 different depths of a SGC model and $MASED$ values in different layers within the model. We show how $MASED$ evolve in the first, the middle and the last layer of each model. We also include the accuracy achieved by each model.}
    \label{fig:mads_sgc}
\end{figure}

\begin{figure}[H]
    \centering
    \includegraphics[width=\linewidth]{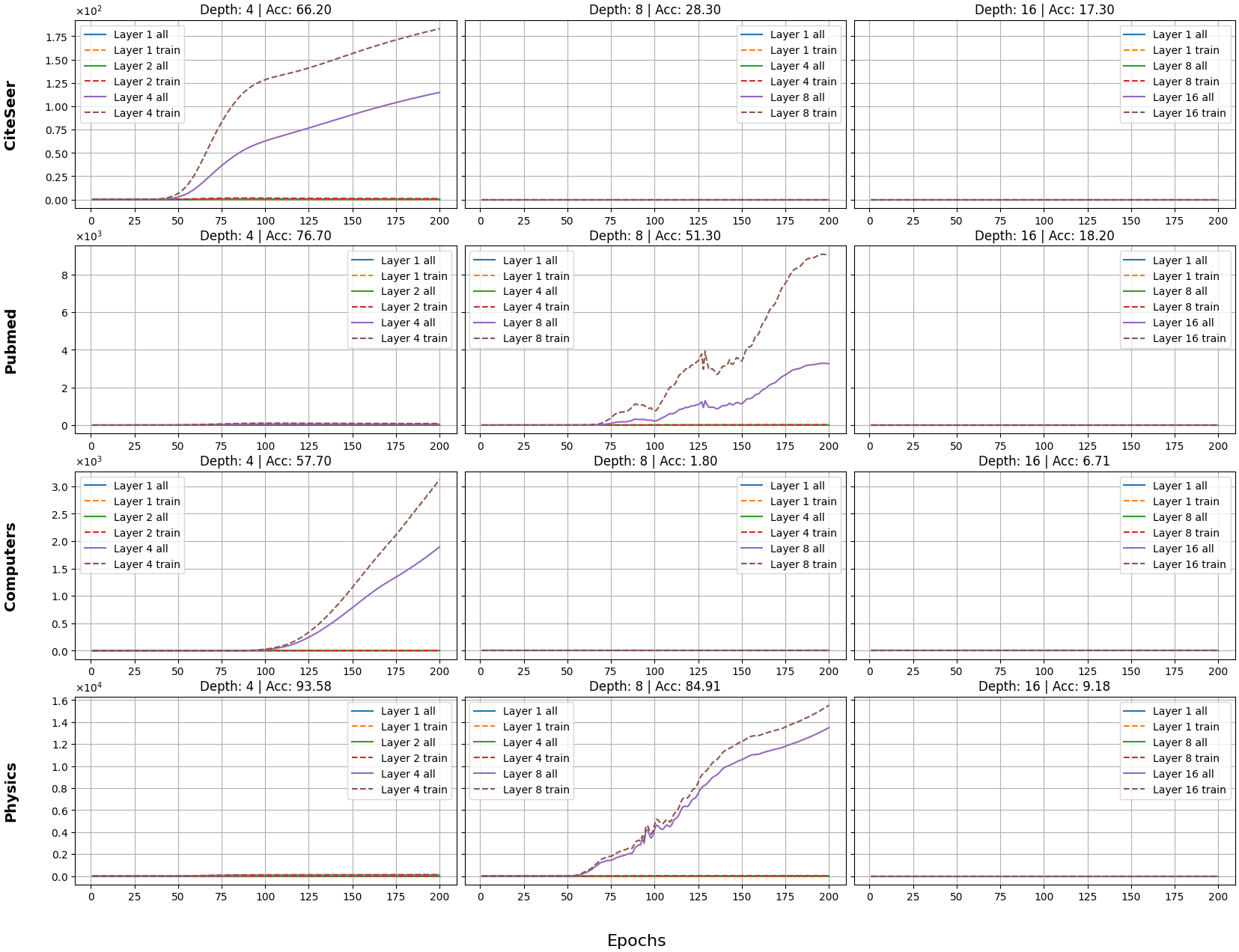}
    \caption{Epoch evolution of the Mean Average Squared Euclidean Distance ($MASED$) value of the embeddings of all nodes and training nodes. We show results for 3 different depths of a GCN model and $MASED$ values in different layers within the model. We show how $MASED$ evolve in the first, the middle and the last layer of each model. We also include the accuracy achieved by each model.}
    \label{fig:mads_gcn_rest}
\end{figure}

\begin{figure}[H]
    \centering
    \includegraphics[width=\linewidth]{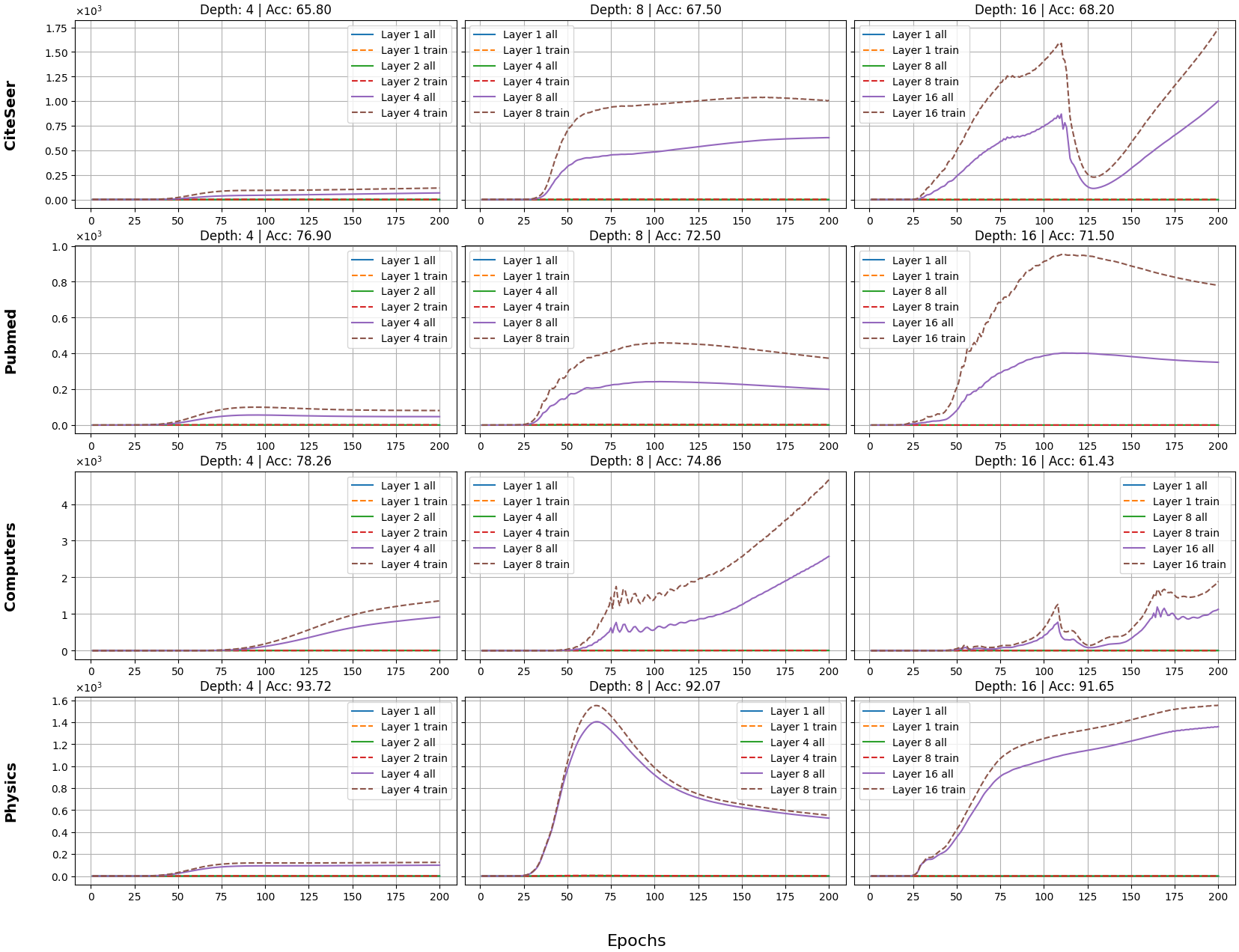}
    \caption{Epoch evolution of the Mean Average Squared Euclidean Distance ($MASED$) value of the embeddings of all nodes and training nodes. We show results for 3 different depths of a ResGCN model and $MASED$ values in different layers within the model. We show how $MASED$ evolve in the first, the middle and the last layer of each model. We also include the accuracy achieved by each model.}
    \label{fig:mads_resgcn_rest}
\end{figure}

\begin{figure}[H]
    \centering
    \includegraphics[width=\linewidth]{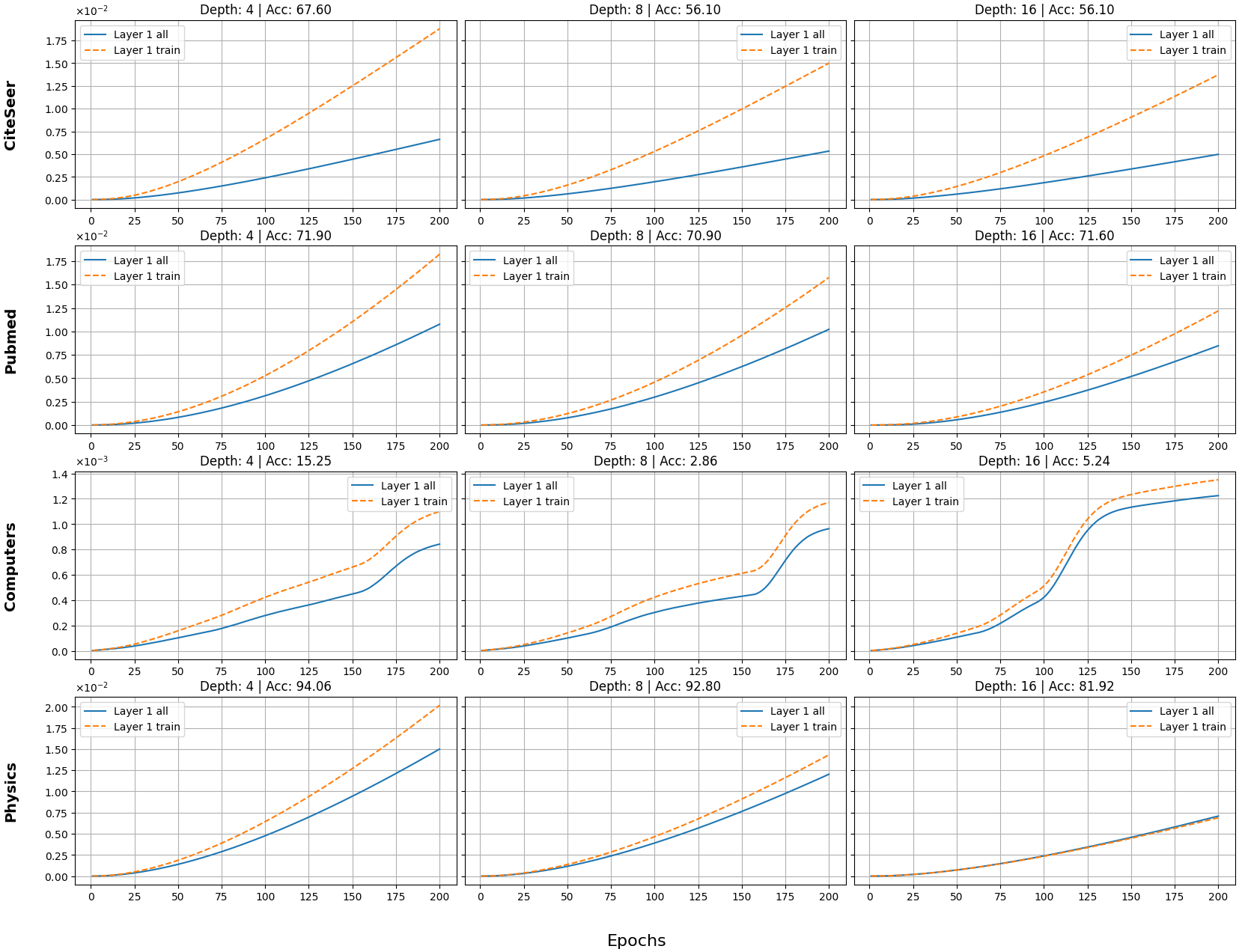}
    \caption{Epoch evolution of the Mean Average Squared Euclidean Distance ($MASED$) value of the embeddings of all nodes and training nodes. We show results for 3 different depths of a SGC model and $MASED$ in different layers within the model. We show how $MASED$ evolve in the first, the middle and the last layer of each model. We also include the accuracy achieved by each model.}
    \label{fig:mads_sgc_rest}
\end{figure}

Figures \ref{fig:mads_resgcn}, and \ref{fig:mads_sgc} present the $MASED$ evolution for ResGCN and SGC models. Additionally, Figures \ref{fig:mads_gcn_rest}, \ref{fig:mads_resgcn_rest}, and \ref{fig:mads_sgc_rest} present the $MASED$ evolution on the \textit{CiteSeer, Pubmed, Computers,} and \textit{Physics} datasets for all models.

\section{Evolution Plots for Embedding Norms \& Centroids Angles}\label{apdx:B}

\begin{figure}
    \centering
    \includegraphics[width=\linewidth]{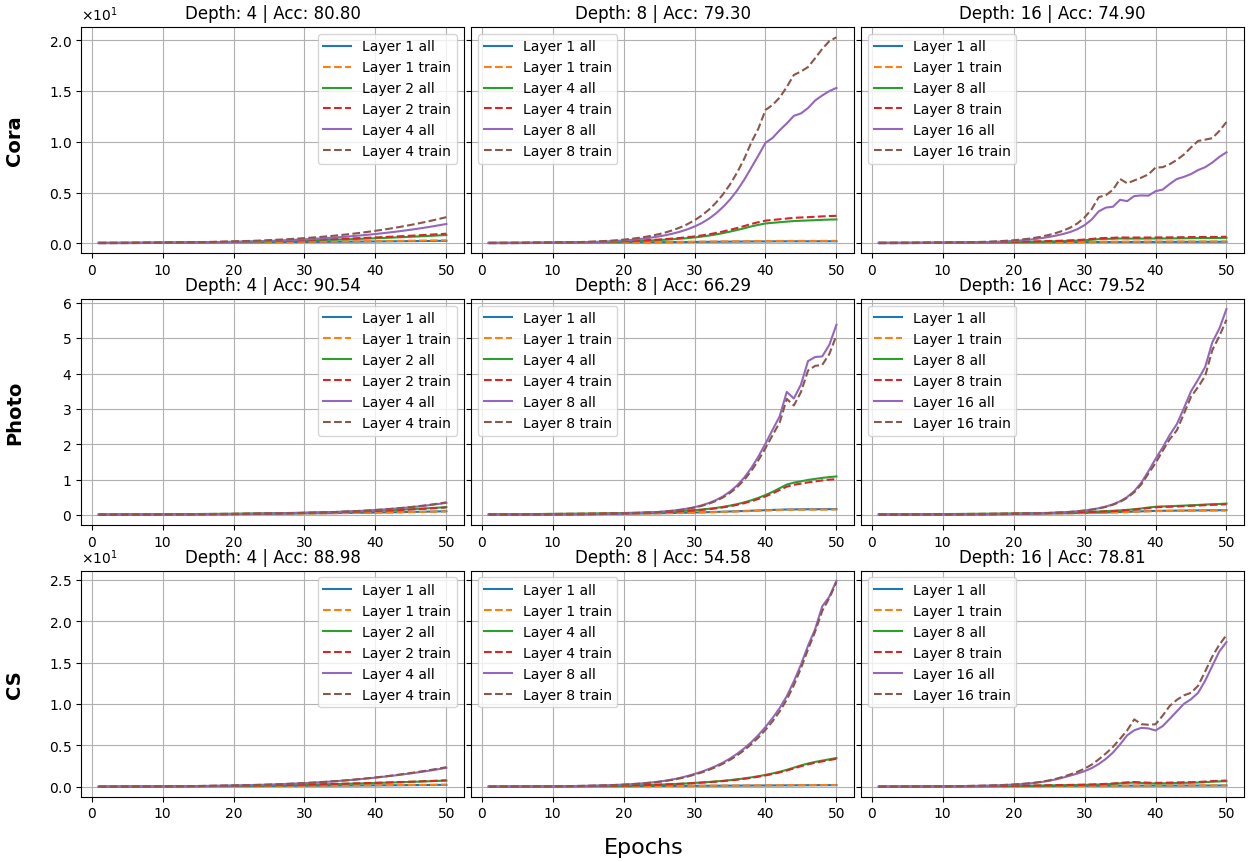}
    \caption{Epoch evolution of the average value of the norms of the embeddings of all nodes and training nodes separately. We show results for 3 different depths of a ResGCN model and average norm values in different layers within the model. We show how norms evolve in the first, the middle and the last layer of each model. We also include the accuracy achieved by each model.}
    \label{fig:norms_resgcn}
\end{figure}

\begin{figure}[H]
    \centering
    \includegraphics[width=\linewidth]{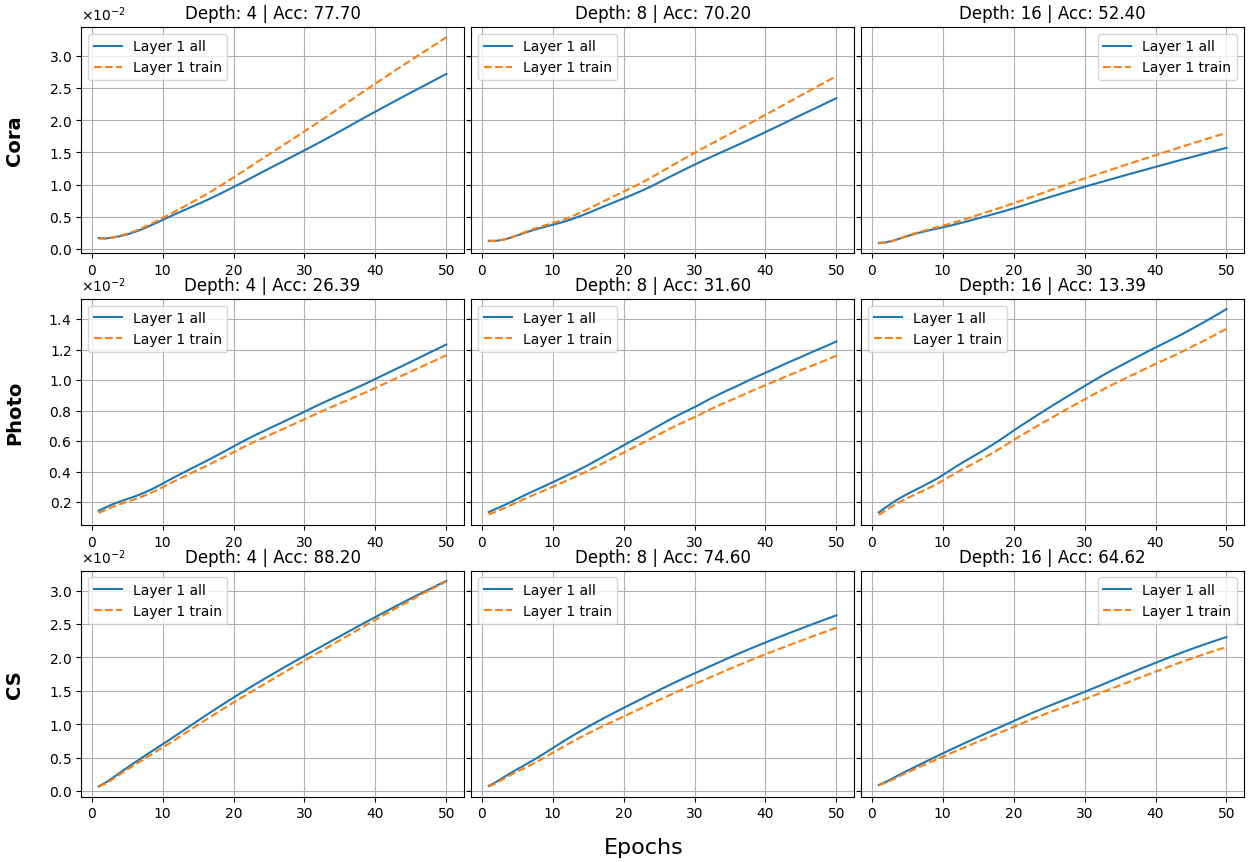}
    \caption{Epoch evolution of the average value of the norms of the embeddings of all nodes and training nodes separately. We show results for 3 different depths of a SGC model and average norm values in different layers within the model. We show how norms evolve in the first, the middle and the last layer of each model. We also include the accuracy achieved by each model.}
    \label{fig:norms_sgc}
\end{figure}

Figure \ref{fig:norms_resgcn} and Figure \ref{fig:norms_sgc} show the evolution of the norms during 50 epochs for ResGCN and SGC models.

\begin{figure}
    \centering
    \includegraphics[width=\linewidth]{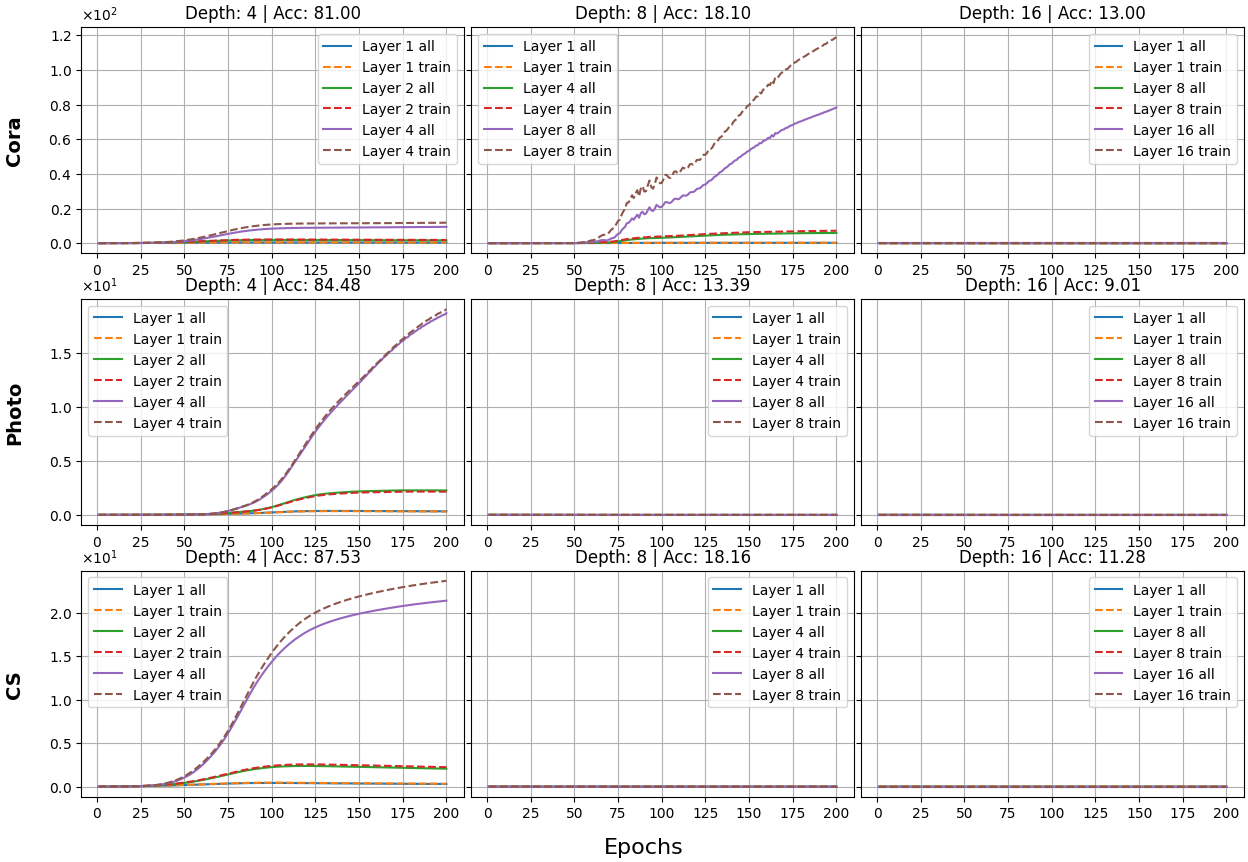}
    \caption{Epoch evolution of the average value of the norms of the embeddings of all nodes and training nodes separately. We show results for 3 different depths of a GCN model and average norm values in different layers within the model. We show how norms evolve in the first, the middle and the last layer of each model. We also include the accuracy achieved by each model.}
    \label{fig:norms_gcn_200}
\end{figure}

\begin{figure}
    \centering
    \includegraphics[width=\linewidth]{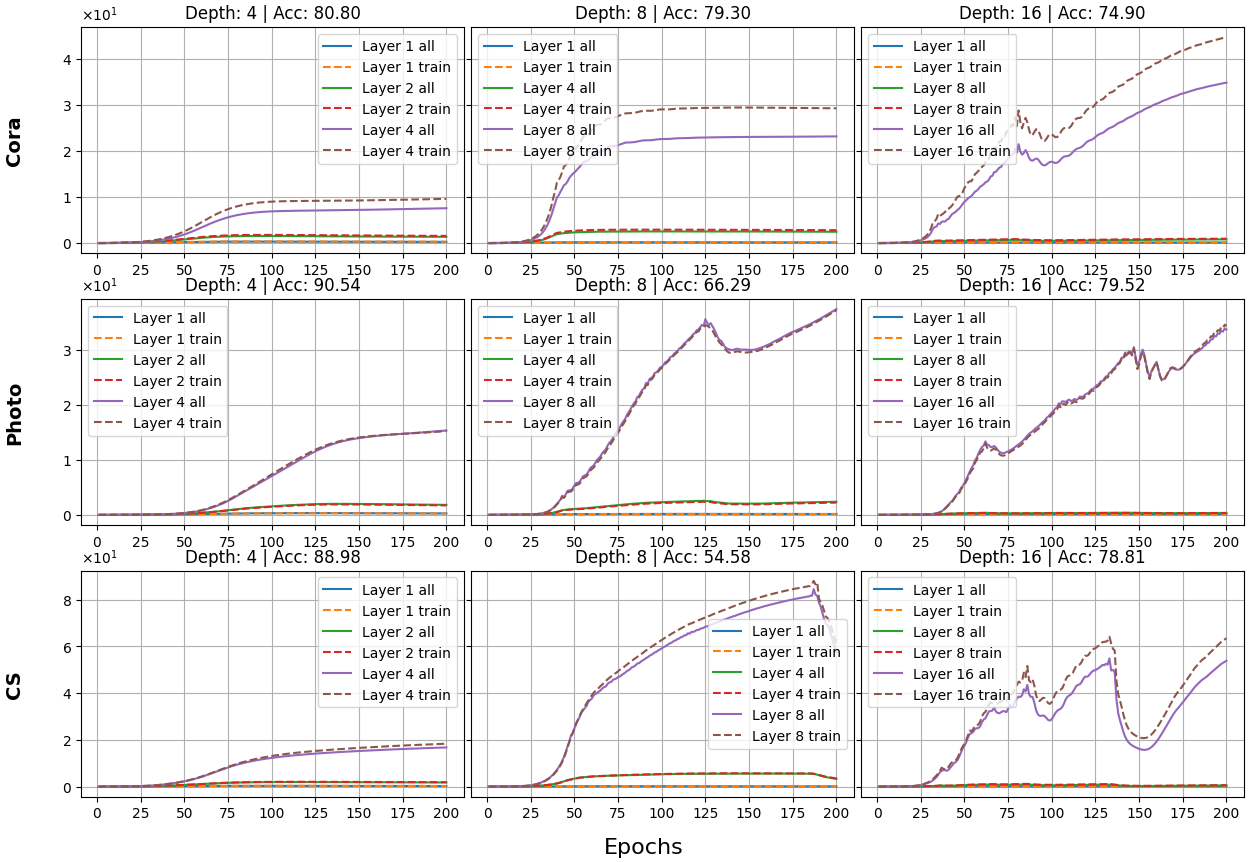}
    \caption{Epoch evolution of the average value of the norms of the embeddings of all nodes and training nodes separately. We show results for 3 different depths of a ResGCN model and average norm values in different layers within the model. We show how norms evolve in the first, the middle and the last layer of each model. We also include the accuracy achieved by each model.}
    \label{fig:norms_resgcn_200}
\end{figure}

\begin{figure}[H]
    \centering
    \includegraphics[width=\linewidth]{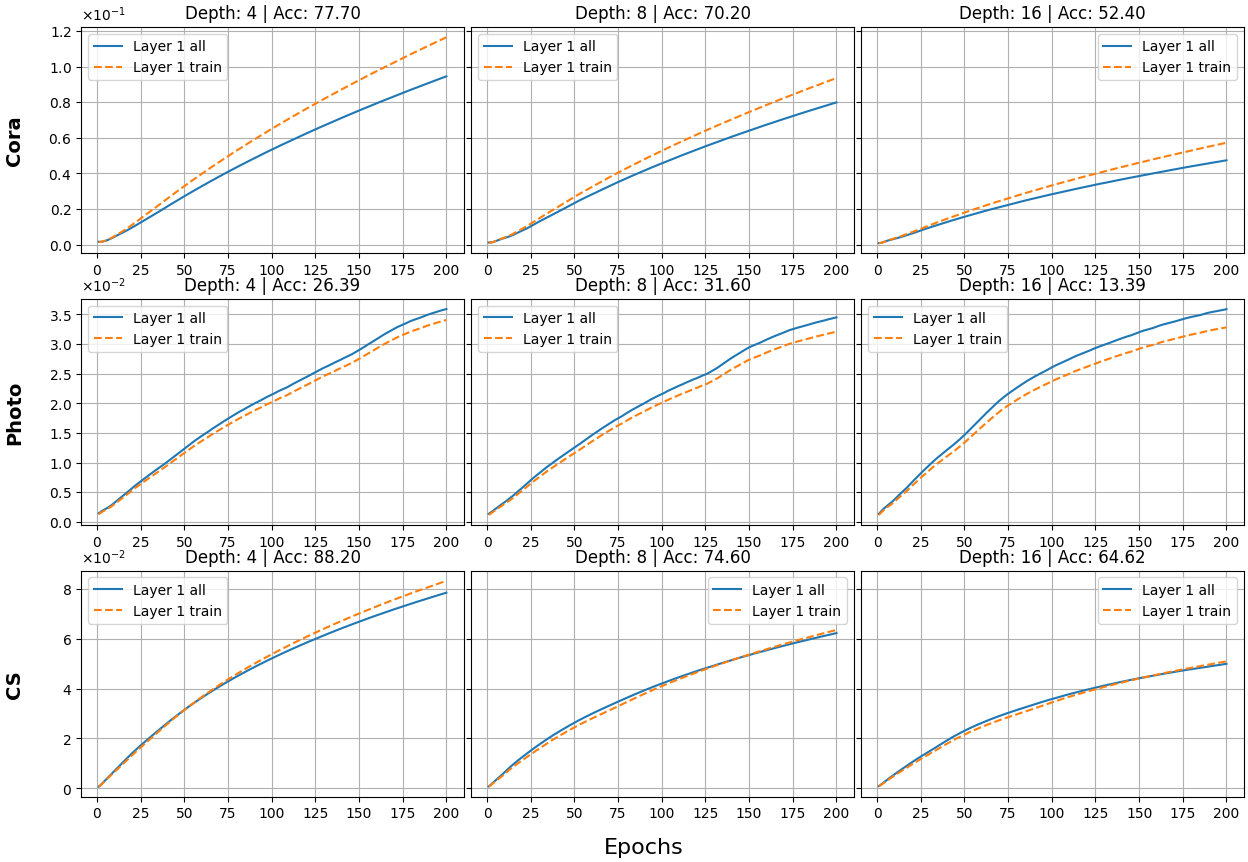}
    \caption{Epoch evolution of the average value of the norms of the embeddings of all nodes and training nodes separately. We show results for 3 different depths of a SGC model and average norm values in different layers within the model. We show how norms evolve in the first, the middle and the last layer of each model. We also include the accuracy achieved by each model.}
    \label{fig:norms_sgc_200}
\end{figure}

\noindent Figures \ref{fig:norms_gcn_200}, \ref{fig:norms_resgcn_200}, and \ref{fig:norms_sgc_200} present the evolution of norms for 200 epochs for GCN, ResGCN, and SGC models.

\begin{figure}[H]
    \centering
    \includegraphics[width=\linewidth]{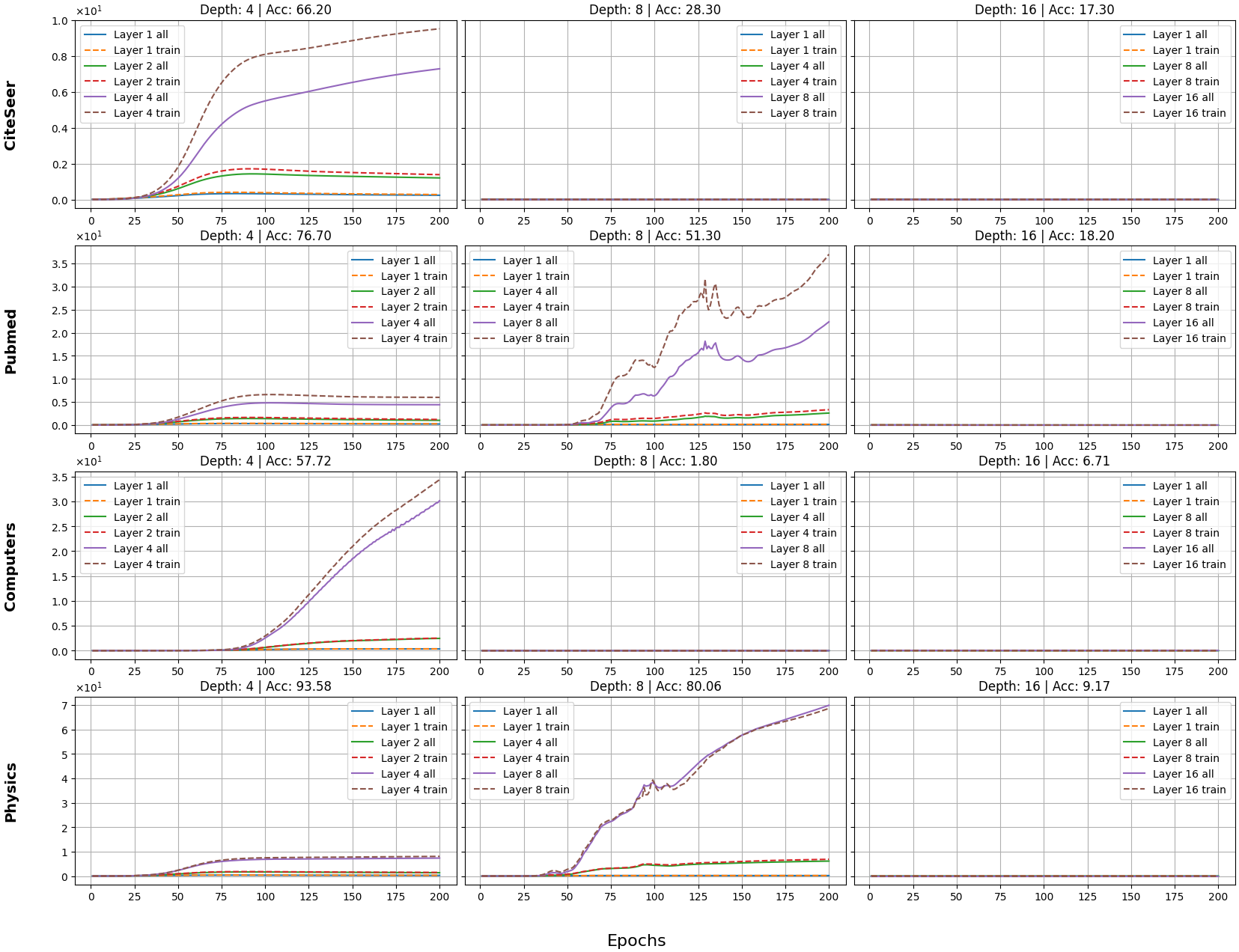}
    \caption{Epoch evolution of the average value of the norms of the embeddings of all nodes and training nodes separately. We show results for 3 different depths of a GCN model and average norm values in different layers within the model. We show how norms evolve in the first, the middle and the last layer of each model. We also include the accuracy achieved by each model.}
    \label{fig:norms_gcn_rest_200}
\end{figure}

\begin{figure}[H]
    \centering
    \includegraphics[width=\linewidth]{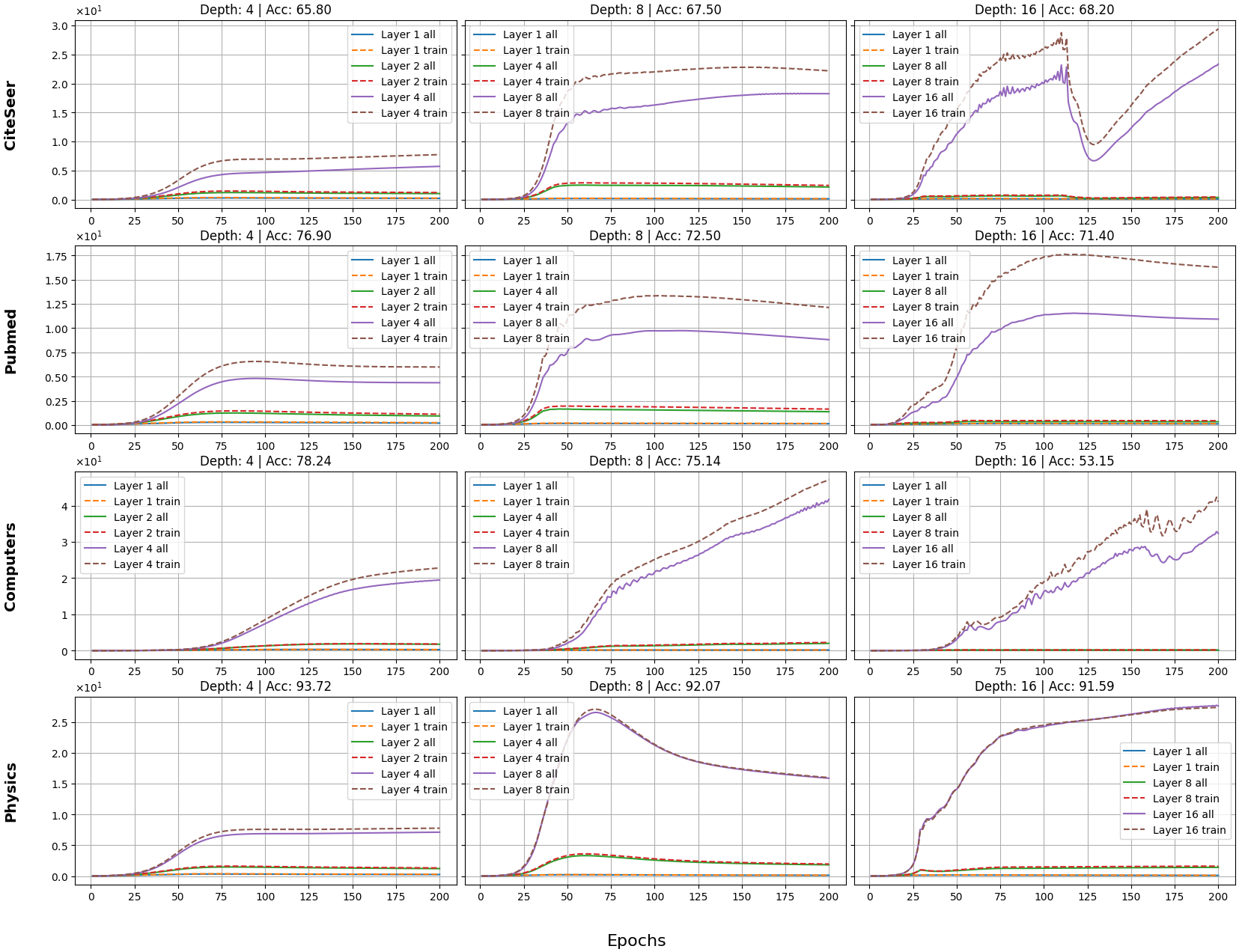}
    \caption{Epoch evolution of the average value of the norms of the embeddings of all nodes and training nodes separately. We show results for 3 different depths of a ResGCN model and average norm values in different layers within the model. We show how norms evolve in the first, the middle and the last layer of each model. We also include the accuracy achieved by each model.}
    \label{fig:norms_resgcn_rest_200}
\end{figure}

\begin{figure}[H]
    \centering
    \includegraphics[width=\linewidth]{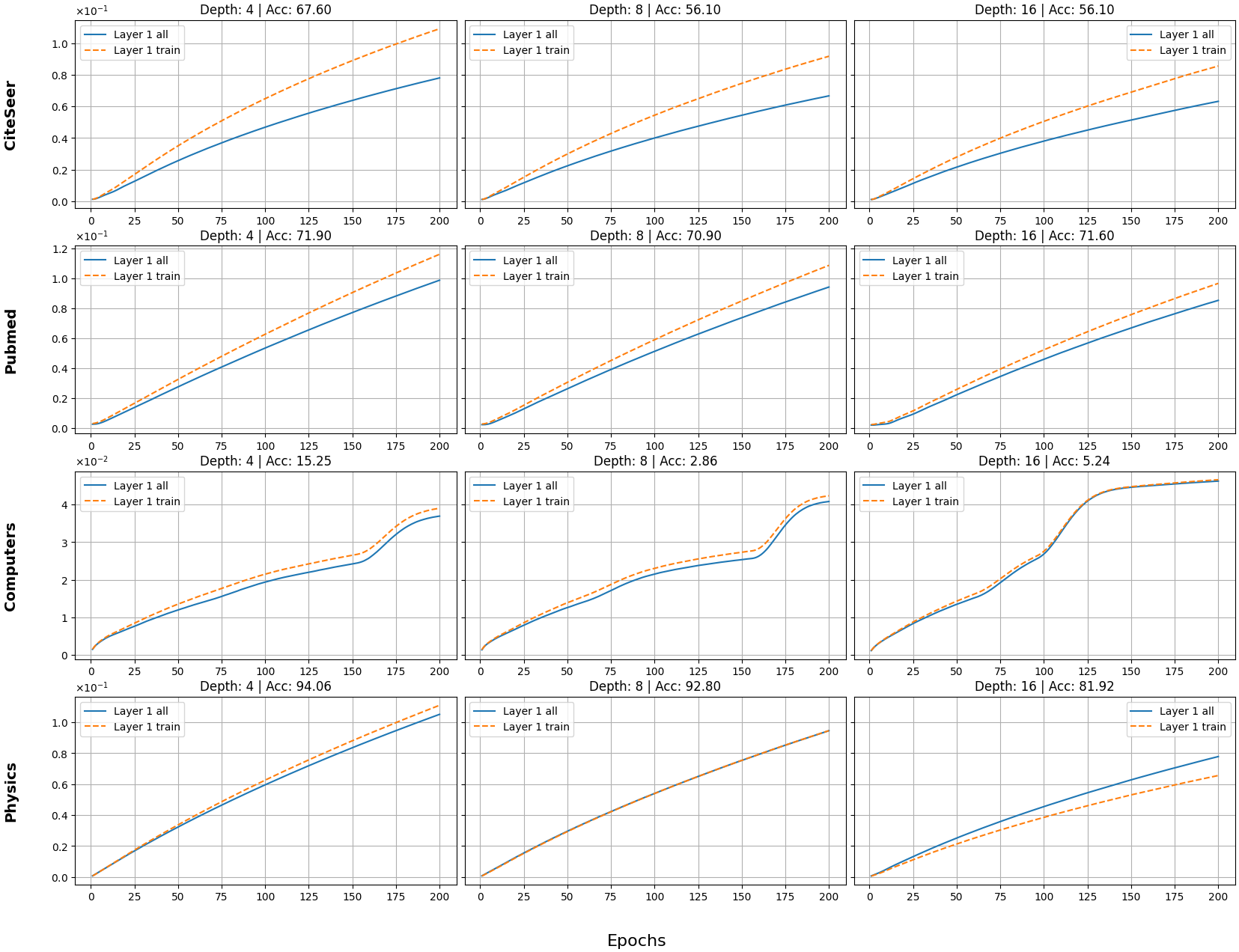}
    \caption{Epoch evolution of the average value of the norms of the embeddings of all nodes and training nodes separately. We show results for 3 different depths of a SGC model and average norm values in different layers within the model. We show how norms evolve in the first, the middle and the last layer of each model. We also include the accuracy achieved by each model.}
    \label{fig:norms_sgc_rest_200}
\end{figure}

\noindent Figures \ref{fig:norms_gcn_rest_200}, \ref{fig:norms_resgcn_rest_200}, and \ref{fig:norms_sgc_rest_200} show the norm evolution on the \textit{CiteSeer, Pubmed, Computers,} and \textit{Physics} datasets for all models under investigation.

\begin{figure}[H]
    \centering
    \includegraphics[width=\linewidth]{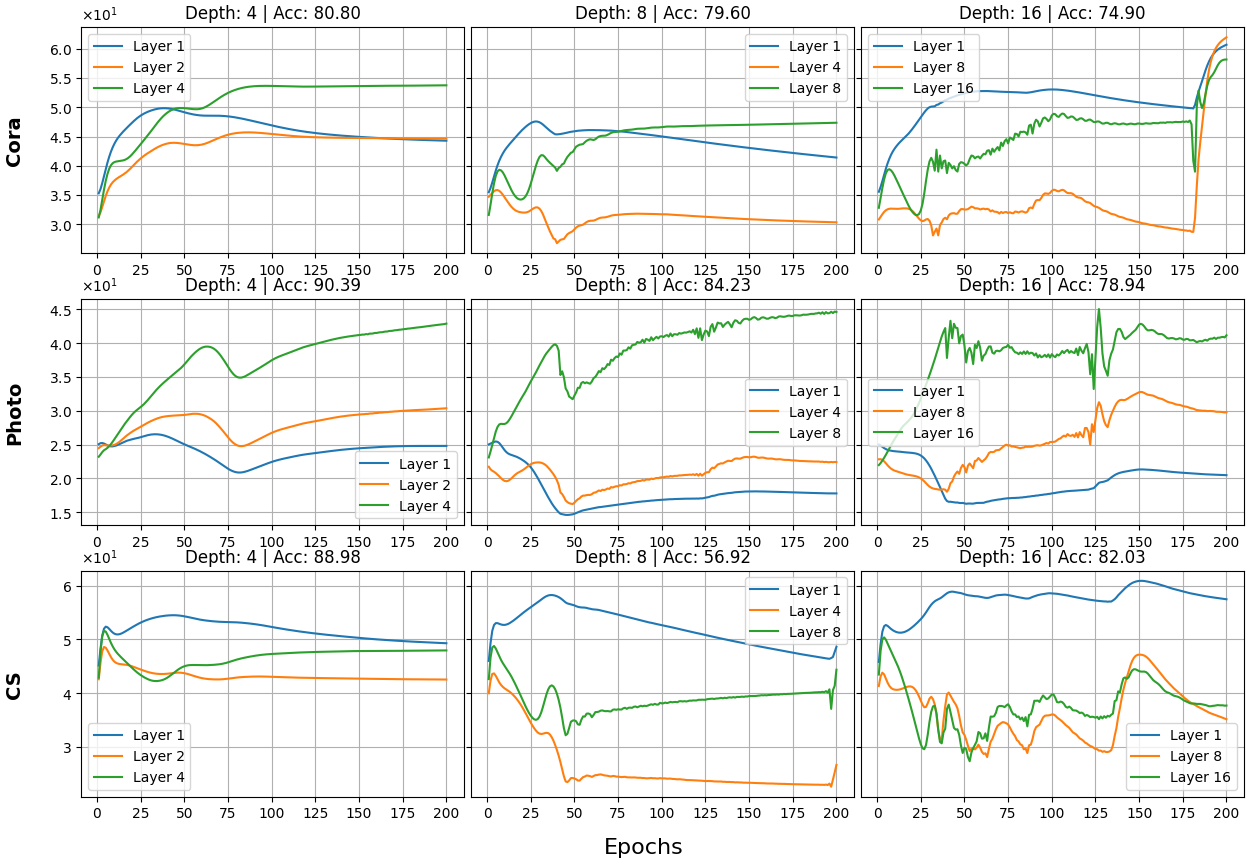}
    \caption{Epoch evolution of the average value of the angles between the centroids of the embeddings of the training nodes. We show results for 3 different depths of a ResGCN model and average norm values in different layers within the model. We show how angles evolve in the first, the middle and the last layer of each model. We also include the accuracy achieved by each model.}
    \label{fig:angles_resgcn}
\end{figure}

\begin{figure}[H]
    \centering
    \includegraphics[width=\linewidth]{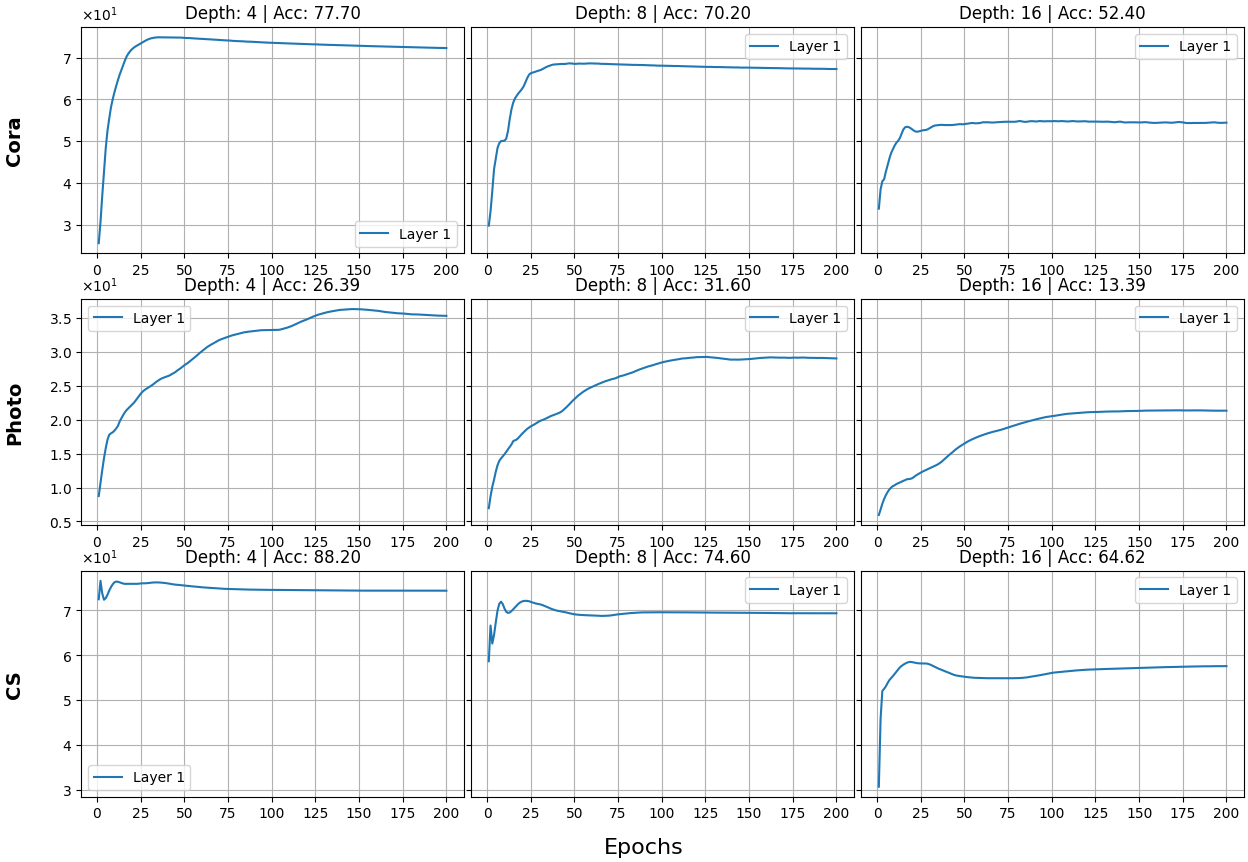}
    \caption{Epoch evolution of the average value of the angles between the centroids of the embeddings of the training nodes. We show results for 3 different depths of a SGC model and average norm values in different layers within the model. We show how angles evolve in the first, the middle and the last layer of each model. We also include the accuracy achieved by each model.}
    \label{fig:angles_sgc}
\end{figure}

\begin{figure}[H]
    \centering
    \includegraphics[width=\linewidth]{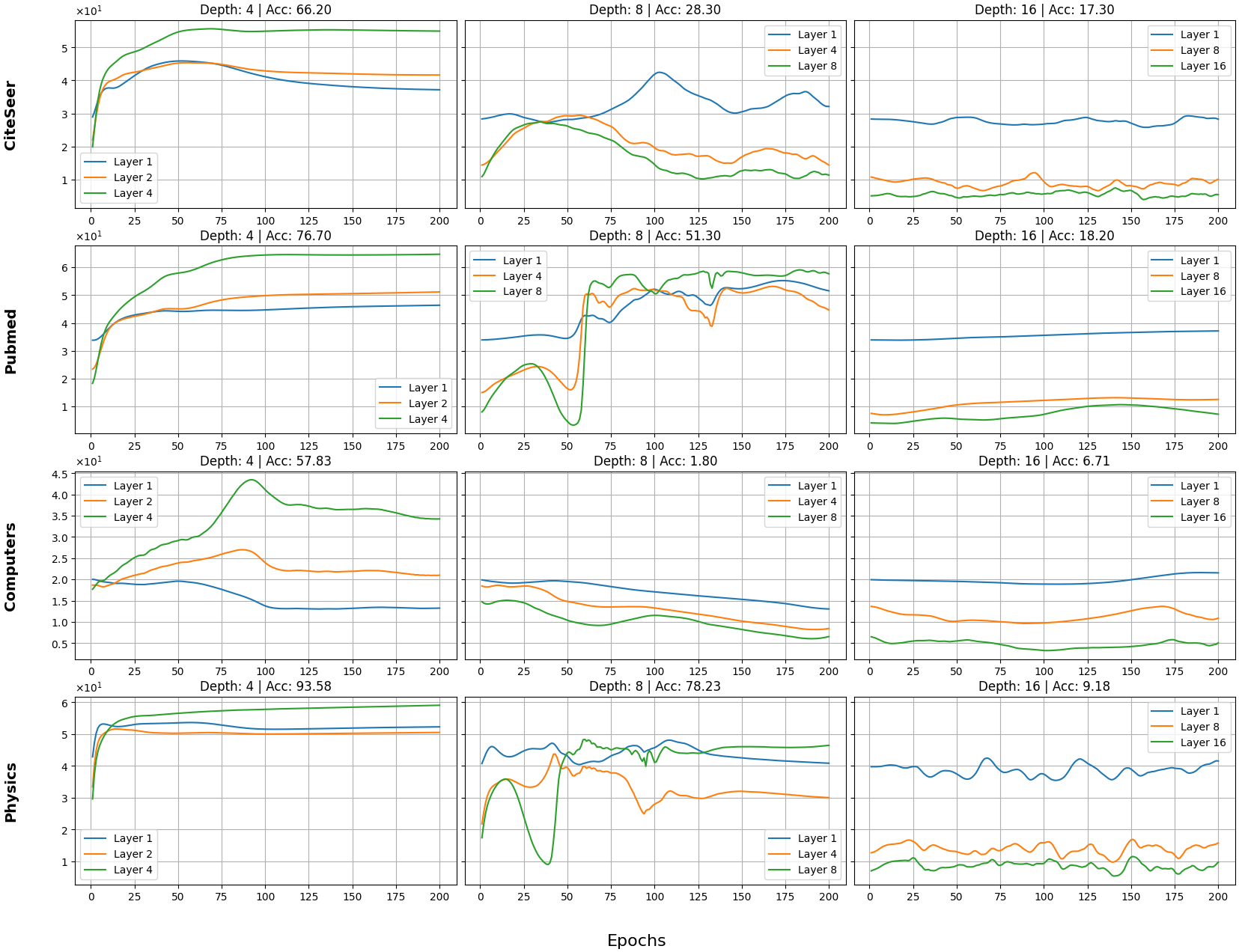}
    \caption{Epoch evolution of the average value of the angles between the centroids of the embeddings of the training nodes. We show results for 3 different depths of a GCN model and average norm values in different layers within the model. We show how angles evolve in the first, the middle and the last layer of each model. We also include the accuracy achieved by each model.}
    \label{fig:angles_gcn_rest}
\end{figure}

\begin{figure}[H]
    \centering
    \includegraphics[width=\linewidth]{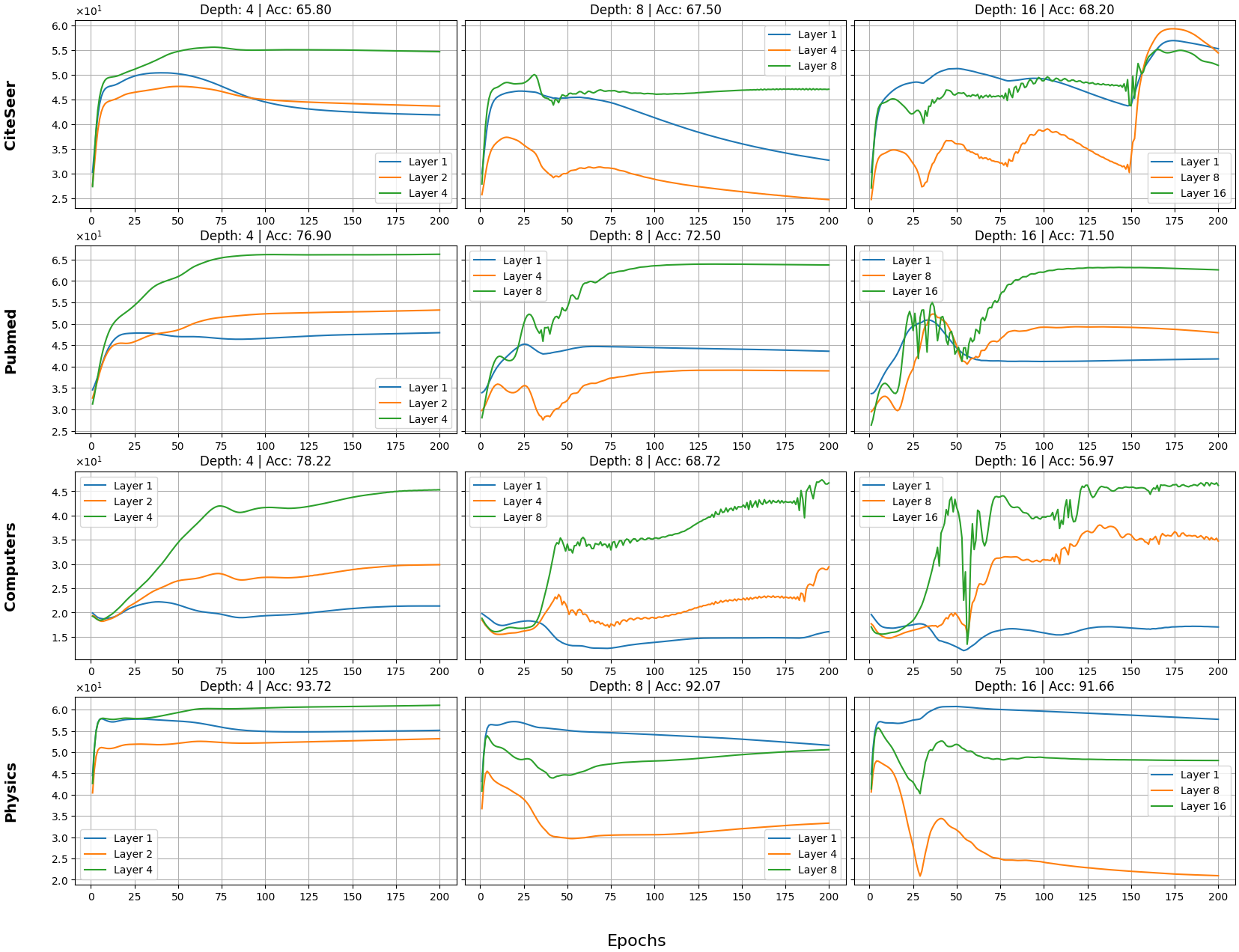}
    \caption{Epoch evolution of the average value of the angles between the centroids of the embeddings of the training nodes. We show results for 3 different depths of a ResGCN model and average norm values in different layers within the model. We show how angles evolve in the first, the middle and the last layer of each model. We also include the accuracy achieved by each model.}
    \label{fig:angles_resgcn_rest}
\end{figure}

\begin{figure}[H]
    \centering
    \includegraphics[width=\linewidth]{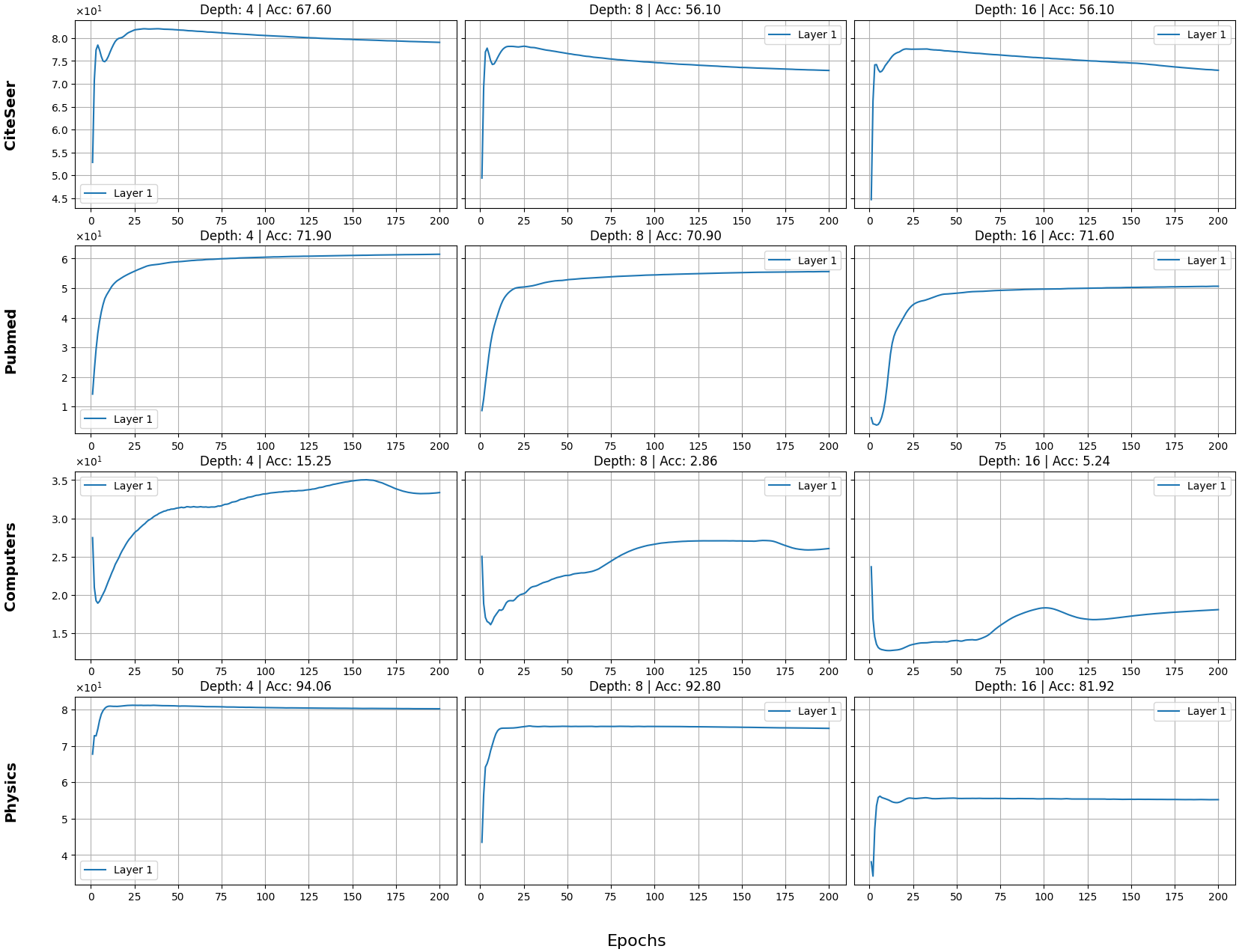}
    \caption{Epoch evolution of the average value of the angles between the centroids of the embeddings of the training nodes. We show results for 3 different depths of a SGC model and average norm values in different layers within the model. We show how angles evolve in the first, the middle and the last layer of each model. We also include the accuracy achieved by each model.}
    \label{fig:angles_sgc_rest}
\end{figure}

\noindent Figures \ref{fig:angles_resgcn}, and \ref{fig:angles_sgc} show the evolution of the average angle between the centroids of the embeddings of the training nodes for ResGCN and SGC. Figures \ref{fig:angles_gcn_rest}, \ref{fig:angles_resgcn_rest}, and \ref{fig:angles_sgc_rest} present the angle evolution on the \textit{CiteSeer, Pubmed, Computers,} and \textit{Physics} datasets.

\section{Regularization plots}\label{apdx:D}

\begin{figure} [H]
    \centering
    \includegraphics[width=\linewidth]{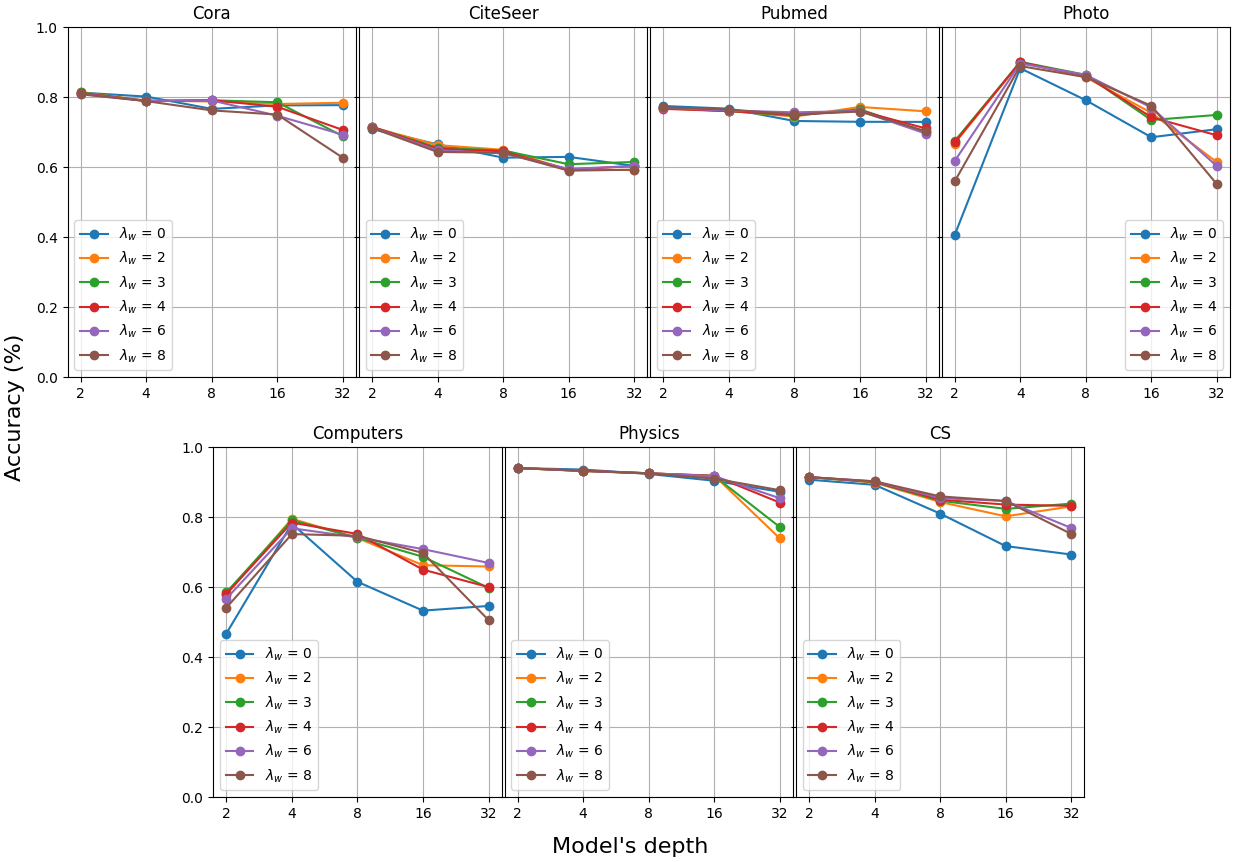}
    \caption{Comparison between a ResGCN with and without the proposed regularization across 7 datasets for varying depth. We include results for different values of $\lambda_w$.}
    \label{fig:benchmarks_resgcn}
\end{figure}

\begin{figure}[H]
    \centering
    \includegraphics[width=\linewidth]{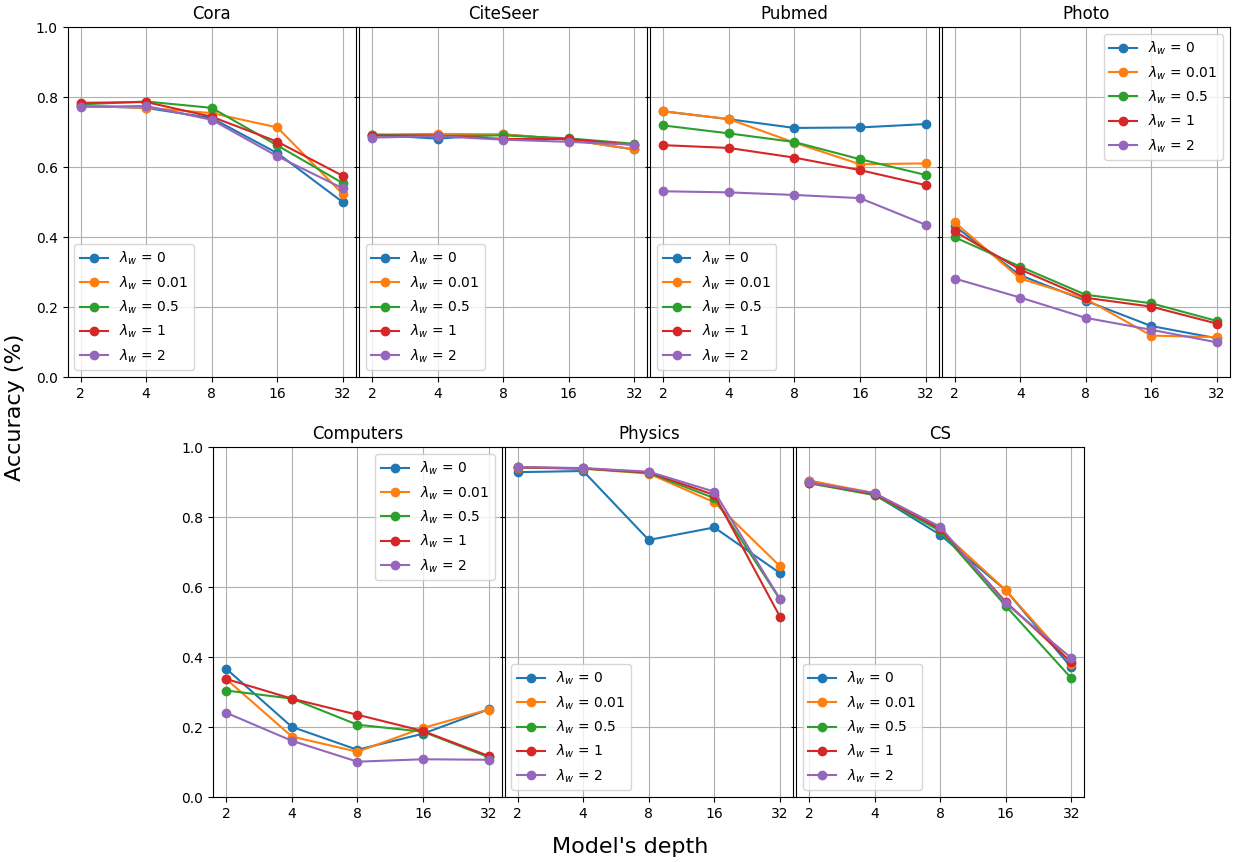}
    \caption{Comparison between a SGC with and without the proposed regularization across 7 datasets for varying depth. We include results for different values of $\lambda_w$.}
    \label{fig:benchmarks_sgc}
\end{figure}

Figures \ref{fig:benchmarks_resgcn}, and \ref{fig:benchmarks_sgc} present the results of ResGCN and SGC on every dataset with and without the proposed regularization.

\section{Plots on variable number of SGC layers}\label{apdx:E}

\begin{figure}[H]
    \centering
    \includegraphics[width=\linewidth]{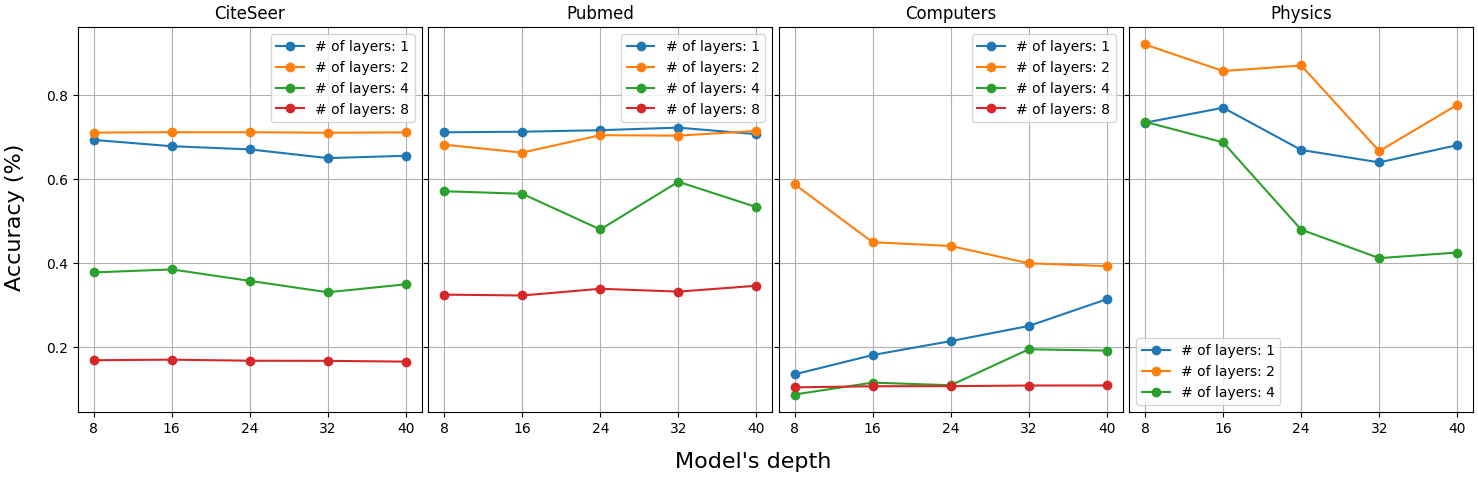}
    \caption{Comparison between SGC models that have different number of SGC layers stacked across 4 different datasets for varying depth. In every depth all models have access to the same information. We only vary the number of trainable weight matrices (i.e., the number of SGC layers).}
    \label{fig:varying_SGCs_rest}
\end{figure}

Figure \ref{fig:varying_SGCs_rest} presents the performance of SGC models varying number of layers on the \textit{CiteSeer, Pubmed, Computers,} and \textit{Physics} datasets. For the \textit{Physics} dataset, we could not train a model with 8 SGC layers, due to hardware limitations.

\section{Dataset Statistics}\label{apdx:G}

\begin{table}[H]
    \centering
    \caption{The statistics of all datasets used in this work.}
    \begin{tabular}{ccccc}
    \hline
    Datasets & \# Nodes & \# Edges & \# Classes & \# Features  \\
     \hline
     Cora & 2708 & 10556 & 7 & 1433\\
     CiteSeer & 3327 & 9104 & 6 & 3703\\
     Pubmed & 19717 & 88648 & 3 & 500\\
     Physics & 34493 & 495924 & 5 & 8415 \\
     CS & 18333 & 163788 & 15 & 6805 \\
     Photo & 7650 & 238162 & 8 & 745\\
     Computers & 13752 & 491722 & 10 & 767\\
     \hline
    \end{tabular}
    \label{tab:my_label}
\end{table}

\section{Hyperparameters}\label{apdx:H}

\begin{table}[h]
    \centering
    \caption{Hyperparameter search space used for finding the optimal configuration.}
    \begin{tabular}{l|c}
        \hline
        \textbf{Hyperparameter} & \textbf{Search Space} \\
        \hline
        Learning Rate ($lr$) & \{1e-4, 6e-4, 1e-3, 6e-3, 1e-2, 6e-2\} \\
        Hidden Dimension & \{64, 128, 256\} \\
        Number of Layers & \{2, 4, 8, 16, 24, 32, 40\} \\
        Weight Decay & \{5e-4, 1e-3, 0\} \\
        Epochs & \{200, 1500, 3000\}\\
        $\lambda_w$ & \{0, 0.01, 0.5, 1, 2, 3, 4, 6, 8, 10, 12\}\\
        \hline
    \end{tabular}
    \label{tab:hyperparams}
\end{table}

\vskip 0.2in
\bibliography{sample}

\end{document}